\documentclass{article}
\usepackage{iclr2027_conference,times}
\usepackage{graphicx}
\usepackage{xcolor}

\usepackage{amsmath,amsfonts,bm}

\def\eqref#1{equation~\ref{#1}}
\def\1{\bm{1}}

\DeclareMathAlphabet{\mathsfit}{\encodingdefault}{\sfdefault}{m}{sl}
\SetMathAlphabet{\mathsfit}{bold}{\encodingdefault}{\sfdefault}{bx}{n}

\newcommand{\E}{\mathbb{E}}

\newcommand{\R}{\mathbb{R}}

\renewcommand{\eqref}[1]{(\ref{#1})}

\usepackage{amssymb}
\usepackage{amsthm}
\usepackage{mathtools}

\usepackage{algorithm}
\usepackage{algorithmic}

\usepackage{booktabs}
\usepackage{array}
\usepackage{multirow}
\usepackage{subcaption}
\usepackage{hyperref}
\usepackage{url}
\usepackage{cleveref}

\crefname{assumption}{Assumption}{Assumptions}
\Crefname{assumption}{Assumption}{Assumptions}

\newtheorem{theorem}{Theorem}[section]
\newtheorem{lemma}[theorem]{Lemma}

\newtheorem{proposition}[theorem]{Proposition}

\theoremstyle{definition}
\newtheorem{definition}[theorem]{Definition}

\theoremstyle{remark}

\DeclareMathOperator{\range}{range}

\DeclareMathOperator{\supp}{supp}

\newcommand{\Vc}{\mathcal{V}}
\newcommand{\Ac}{\mathcal{A}}
\newcommand{\Vsc}{\Vc_{S_c}}
\newcommand{\Lsage}{\mathcal{L}_{\mathrm{sage}}}
\newcommand{\Lfm}{\mathcal{L}_{\mathrm{FM}}}
\newcommand{\Droute}{D_{\mathrm{route}}}
\newcommand{\ksplit}{k_{\mathrm{split}}}

\newcommand{\vcfg}{v_{\mathrm{cfg}}}
\newcommand{\videal}{v_{\mathrm{ideal}}}

\title{\LARGE SAGE: Subspace Alignment for Classifier-Free Guidance in Mixture-of-Experts Diffusion Models}

\iclrfinalcopy

\newcommand{\appref}[1]{Appendix~\ref*{#1}}

\newcommand{\suppcref}[1]{\cref*{#1}}

\newcommand{\maincref}[1]{\cref*{#1} of the main paper}
\newcommand{\maineqref}[1]{Eq.~(\ref*{#1}) of the main paper}

\begin{document}
\maketitle

\begin{center}
    \vspace{-1.8cm}
    \large \textbf{Boyu Zhang$^{1}$,  Yangming Cheng$^{1}$, Ning Zhang$^{1}$, Pengfei Liu$^{1}$, \\ Weijie Li$^{1}$, Yifan Gao$^{1}$, Hangyu Li$^{1}$, Litong Gong$^{1,*}$} \\
    \vspace{0.5em}
    \normalsize
    $^1$Alibaba Token Hub, Alibaba Group \\
    \vspace{0.8em}
\end{center}

\begingroup
\renewcommand{\thefootnote}{\fnsymbol{footnote}}
\footnotetext[1]{Corresponding author.}
\endgroup

%% ====================================================================
%% ABSTRACT
%% ====================================================================
\begin{abstract}
Diffusion Transformers with Mixture-of-Experts (MoE) routing are a leading recipe for scaling generative models. Classifier-Free Guidance (CFG) is essential for generation quality, yet excessively high guidance scales trigger collapse. We identify a previously unreported failure mode in their combination: the two CFG branches route independently, so their realized activations occupy different subspaces. The unconditional write then leaves the conditional subspace, and CFG amplifies that residual linearly in the guidance scale. We propose SAGE, a training-time regularizer that aligns unconditional MoE activations to the conditional subspace without restricting routing diversity, at zero inference cost. Toy experiments show that SAGE dramatically suppresses extreme drift by $9.2\times$. When scaled to a 1B-parameter text-to-image model, SAGE significantly improves generation quality, delivering a $9.3\%$ boost in peak DPG-Bench performance. Extensive experiments demonstrate that SAGE consistently outperforms the baseline.
\end{abstract}

%% ====================================================================
%% 1. INTRODUCTION
%% ====================================================================
\section{Introduction}
\label{sec:intro}

Diffusion Transformers (DiTs)~\citep{peebles2023dit} and their flow-matching
counterparts~\citep{lipman2022fm,liu2022rectified} have become the dominant
paradigm for high-fidelity generation~\citep{seedance2026seedance20advancingvideo,happyhorse26}. Following the success of sparse
experts in language modeling~\citep{fedus2021switch,lepikhin2020gshard},
recent work scales DiTs via Mixture-of-Experts (MoE) feed-forward
blocks, including DiT-MoE~\citep{fei2024scaling}, EC-DiT~\citep{ecdit2025},
Race-DiT~\citep{expertrace2025}, and
others~\citep{diffmoe2025,dense2moe2025}. In parallel, Classifier-Free
Guidance (CFG)~\citep{ho2022cfg} remains the indispensable inference knob
that trades diversity for fidelity by linearly combining conditional and
unconditional velocity fields. Virtually every modern text-to-image system
ships with CFG enabled, and practitioners routinely tune the guidance scale
$s$ to reach the operating point they want. These two ingredients, sparse
experts for capacity and CFG for controllability, are therefore combined in
almost every large-scale generative model being built today.

Despite the popularity of both MoE scaling and CFG conditioning, we
find that their combination harbors a previously overlooked pathology.
When CFG is applied to an MoE-DiT, sample quality degrades far more
severely than in a parameter-matched dense model as the guidance scale
increases. Because the conditional and unconditional forward passes
produce different hidden states, the routers select different
expert subsets for the two branches, and the CFG linear combination
mixes outputs from incompatible subspaces, injecting a spurious
error that grows with the guidance scale. This routing divergence is
intrinsic to MoE and vanishes identically in dense architectures.

We trace this failure to a geometric mechanism.
Different hidden states $h_c\neq h_u$ cause the Top-$k$ routers to
select different expert subsets $S_c\neq S_u$. Write $\Droute$ for this
routing misalignment, and let $\Vsc$ be the subspace spanned by the
realized conditional activations. CFG then injects a leak $\delta(s)$
outside $\Vsc$ that grows with the guidance scale $s$. In a dense model,
$\Droute\equiv 0$, so the failure is MoE-intrinsic.

The mechanism suggests an obvious fix: forcing the two branches onto
the same routes, for instance by penalizing the divergence between
their routing distributions. We find this approach to be fundamentally
flawed. It removes the mismatch, but it also eliminates the
specialization that motivates MoE. Experiments confirm that
constraining routing freedom degrades performance.

Our starting point is that CFG is closed in any common subspace: if both writes lie in a span
$\Vc$, then so does $\vcfg$, regardless of which experts produced them.
It is therefore enough to align the unconditional activations to the
conditional subspace $\Vsc$, while leaving the routers free. This
observation motivates \textbf{SAGE}, \textbf{S}ubspace \textbf{A}lignment for
CF\textbf{G} in Mo\textbf{E} Diffusion Models, a training-time
regularizer that penalizes only the component of the unconditional MoE
output that lies outside the stacked conditional
activations. SAGE requires no architectural change, adds one loss
during training, preserves routing diversity, and incurs zero inference
cost. 

Empirically, the gains are substantial. In toy experiments, SAGE cuts the
worst-case drift at high guidance by $9.2{\times}$ and largely restores mode
purity where the baseline has collapsed. We further scale SAGE up to a 1B-parameter MoE-DiT on the text-to-image task, where it improves the peak DPG-Bench score by $9.3\%$ and keeps that quality as guidance grows. The same pattern holds on GenEval,
where SAGE lifts the peak score by $2.7$\,pp, and the qualitative
comparisons show coherent structure at high guidance where the baseline has
already broken down.

We summarize our contributions as follows:
\begin{itemize}
\item 
  We uncover a novel failure mode specific to MoE-DiTs, termed routing-induced subspace leakage: routing misalignment forces conditional and unconditional branches into disjoint subspaces, which CFG subsequently amplifies linearly with the guidance scale. We show this structural leak is unique to sparse models and vanishes in dense architectures.
\item 
  We propose SAGE, a training-time regularizer that aligns unconditional activations back into the conditional subspace while maintaining expert routing diversity and incurring zero inference overhead. Crucially, we demonstrate that the intuitive alternative of forcing hard routing agreement is markedly sub-optimal.
\item 
  Across extensive experiments on large-scale text-to-image MoE-DiTs, SAGE consistently outperforms baseline models across various CFG scales. Controlled toy experiments and ablation studies validate the underlying mechanism and design choices of SAGE.
\end{itemize}

\vspace{-0.2cm}

%% ====================================================================
%% 2. RELATED WORK
%% ====================================================================
\section{Related Work}
\label{sec:related}

\subsection{Mixture-of-Experts in Diffusion Models}
\label{sec:related-moe}

Sparse mixture-of-experts (MoE) architectures~\citep{lepikhin2020gshard,fedus2021switch} have emerged as a dominant recipe for
scaling diffusion transformers~\citep{peebles2023dit} beyond the capacity of dense feed-forward networks (FFNs).
RAPHAEL~\citep{xue2023raphael} pioneered the use of space-MoE and time-MoE
layers for text-to-image diffusion, demonstrating that routing along spatial
and temporal axes yields strong artistic generation. DiT-MoE~\citep{fei2024scaling} scales sparse DiTs to 16.5B parameters and achieves state-of-the-art performance. EC-DiT~\citep{ecdit2025} replaces token-choice routing with expert-choice
routing~\citep{zhou2022expertchoice} to allocate compute adaptively across image patches, while
Diff-MoE~\citep{diffmoe2025} and Race-DiT~\citep{expertrace2025} further
explore time-aware and space-adaptive expert selection and flexible per-token expert counts.
Dense2MoE~\citep{dense2moe2025} takes a post-hoc route, restructuring
pre-trained FLUX-style~\citep{bfl2024flux} dense DiTs into sparse MoE backbones with 60\% fewer
activated parameters. Almost all of these works evaluate their models
with classifier-free guidance~\citep{ho2022cfg}, yet none of them study how MoE
routing interacts with the dual conditional/unconditional forward passes
that CFG requires.
Most recently, ProMoE~\citep{promoe2025} partitions tokens into
conditional and unconditional subsets via a hard first-stage router and
assigning dedicated experts to the unconditional branch. While this design
provides partial relief analogous to that offered by the shared-expert mechanism
analyzed in \cref{sec:method-sage}, it does not align the realized
activations of the remaining routed experts and sacrifices one expert's
capacity for conditional generation.

\subsection{Classifier-Free Guidance}
\label{sec:related-cfg}

Classifier-free guidance has become the mainstream inference-time mechanism for conditional diffusion, as it improves generation results and enables better semantic control. It relies on jointly trained
conditional and unconditional predictions, which are combined as $(1{+}s)\,v_c - s\,v_u$ at inference. Improving CFG has motivated a line of corrective methods~\citep{hong2023sag,zheng2024characteristic,ahn2024pag,kynkaanniemi2024interval,sadat2025icg}.
GLIDE~\citep{nichol2022glide} empirically compared CFG to CLIP guidance~\citep{radford2021clip} for
text-to-image synthesis, and subsequent work has proposed CFG rescaling,
dynamic schedules, and spatial-inconsistency analyses to mitigate artifacts
at high $s$~\citep{spatialcfg2024}. CADS~\citep{sadat2024cads} anneals
the conditioning to restore diversity, APG~\citep{sadat2025apg} decomposes
the guidance term to remove the oversaturation and artifacts that appear at
large scales, and CFG-Zero*~\citep{fan2025cfgzero} rescales and zero-initializes the
guidance to correct
inaccurate velocity estimates. 
CFG++~\citep{chung2025cfgpp} attributes guidance failures to
iterates drifting off the data manifold and addresses this by
reformulating CFG as a manifold-constrained inverse problem solved at
sampling time.
Another line of work replaces the null-condition branch altogether: PAG~\citep{ahn2024pag}
perturbs self-attention to synthesize a deliberately degraded prediction as the
guidance baseline, in the spirit of self-attention guidance~\citep{hong2023sag}, and
\citet{karras2024autoguidance} propose autoguidance, which improves performance by additionally training an auxiliary model. \citet{cfg_mechanism2025}
investigates how guidance amplifies the top singular directions of the conditional score.
Our analysis complements theirs by revealing a qualitatively different failure mode unique to sparse architectures: routing misalignment places the two CFG branches in different realized subspaces, and CFG then scales the residual of the unconditional write outside the conditional subspace. Unlike autoguidance or guidance-distillation approaches~\citep{meng2023distillcfg,agd2025}, which modify or bypass the unconditional branch, we resolve this structural mismatch during training via an architecture-aware subspace constraint that incurs zero inference cost.
\vspace{-0.1cm}
%% ====================================================================
%% 3. SUBSPACE ALIGNMENT FOR CFG (merged Preliminaries + Method)
%% ====================================================================
\section{SAGE: Subspace Alignment for Classifier-Free Guidance}
\label{sec:method}

\subsection{Preliminaries}
\label{sec:prelim}

\paragraph{Flow Matching.}
Flow matching~\citep{lipman2022fm,liu2022rectified} learns a velocity field
$v_\theta(x,t,c):\R^d\times[0,1]\times\mathcal{C}\to\R^d$ transporting
$\mathcal{N}(0,I_d)$ to $p_{\mathrm{data}}(\cdot\mid c)$. With the flow interpolant $x_t = t\,x_1 + (1{-}t)\,\varepsilon$, the loss is
\begin{equation}
\Lfm(\theta) = \E_{x_1,\varepsilon,t,c}
\Big[\big\lVert v_\theta(x_t,t,c) - (x_1-\varepsilon)\big\rVert_2^2\Big].
\label{eq:fm-loss}
\end{equation}

\paragraph{Classifier-Free Guidance.}
Classifier-free guidance~\citep{ho2022cfg} uses a single model to handle both conditional and unconditional regimes by
replacing $c$ with a null token $\varnothing$ at random during training. At
inference, the two branches are combined as
\begin{equation}
\vcfg(x,t,c,s) = (1+s)\,v_\theta(x,t,c) - s\,v_\theta(x,t,\varnothing).
\label{eq:cfg}
\end{equation}

\paragraph{MoE Layer.}
An MoE layer~\citep{lepikhin2020gshard,fedus2021switch} replaces the FFN with $N$ parallel experts and a sparse
router. Given a hidden state $h\in\R^{d_h}$, the router selects a top-$k$
active set $S(h)\subseteq\{1,\dots,N\}$ with gating weights $g_i(h)$, and
each expert $E_i$ is a two-layer FFN with output projection
$W_2^{(i)}\in\R^{d_h\times d_f}$:
\begin{equation}
f_{\mathrm{MoE}}(h) = \sum_{i\in S(h)} g_i(h)\,E_i(h).
\label{eq:moe}
\end{equation}

\subsection{Routing Misalignment and Subspace Alignment}
\label{sec:method-problem}
\label{sec:method-sage}

\begin{figure}[t]
  \vspace{-0.3cm}
  \centering
  \hspace*{-0.5cm}
  \includegraphics[width=1.04\linewidth]{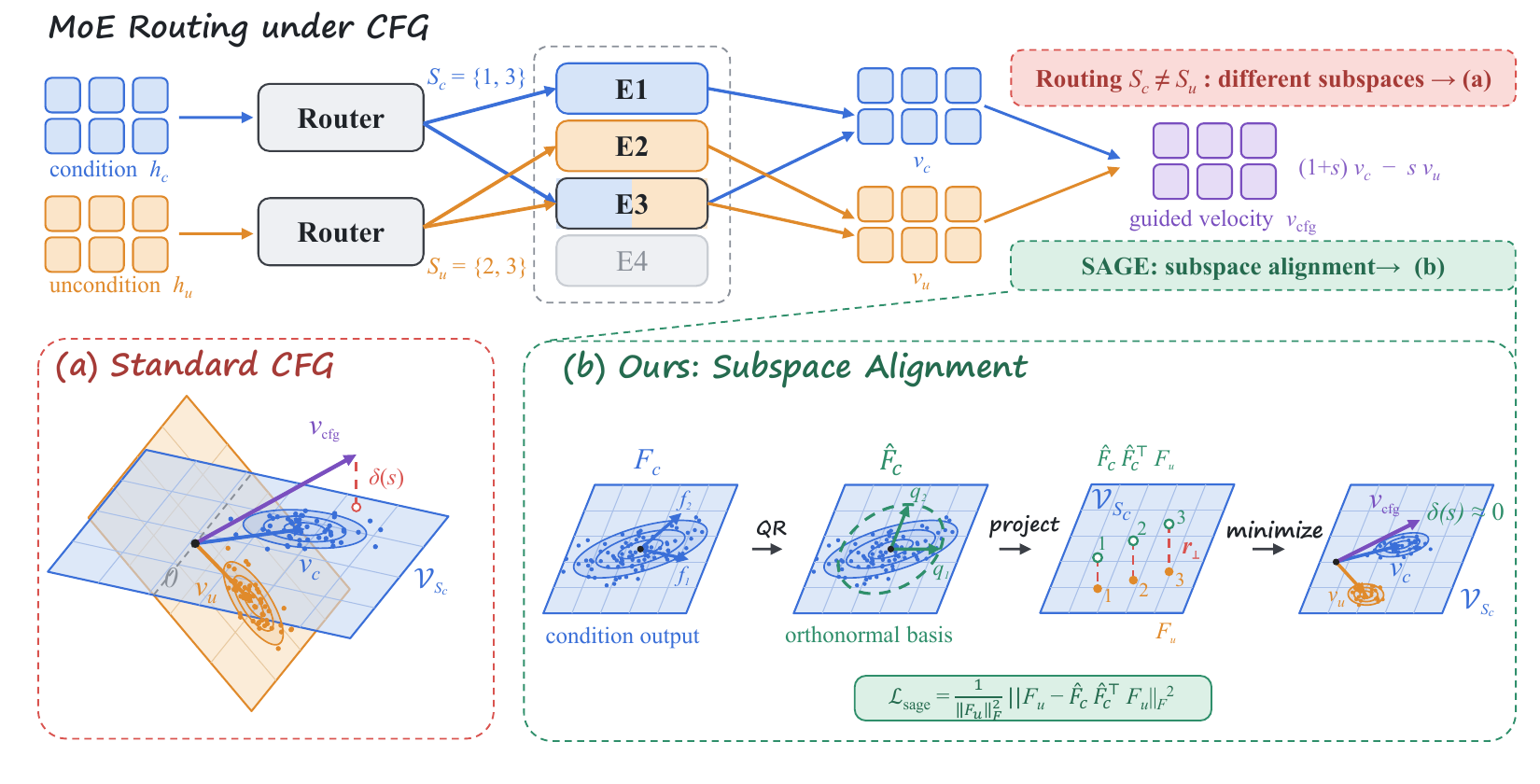}
  \vspace{-0.3cm}
  \caption{\textbf{Subspace alignment.}
  \textbf{Top:} Conditional and unconditional hidden states $h_c$ and $h_u$ are routed independently, so the realized activations occupy different subspaces.
  \textbf{(a)~}Standard CFG injects a leak $\delta(s)$.
  \textbf{(b)~}SAGE uses thin QR of the stacked conditional outputs $F_c$ to yield an orthonormal basis $Q_c$ of $\Vc_{S_c}$. Projecting $F_u$ onto this subspace and penalizing the residual with $\Lsage$ enforces $v_u \in \Vc_{S_c}$ at training time.}
  \label{fig:main}
  \vspace{-0.3cm}
\end{figure}

We write $v_c=f_{\mathrm{MoE}}(h_c)$ and $v_u=f_{\mathrm{MoE}}(h_u)$
for the MoE outputs under condition $c$ and the null token $\varnothing$,
respectively, with active sets $S_c$ and $S_u$. Because the conditioning signal is mixed into earlier layers,
$h_c \neq h_u$, so the routers select different experts:
$S_c \neq S_u$ in general. We quantify this with
\begin{equation}
\Droute := 1 - \frac{|S_c\cap S_u|}{k} \in [0,1].
\label{eq:droute}
\end{equation}

As illustrated in \cref{fig:main} (a), $v_c$ and $v_u$ live in different
subspaces, so the CFG combination~\eqref{eq:cfg} mixes incompatible
directions. Decomposing this combination into shared and exclusive parts gives:
\begin{align}
\vcfg ={}& \underbrace{\sum_{i\in S_c\cap S_u}\!\!\big[(1{+}s)\,g_i^c E_i(h_c)-s\,g_i^u E_i(h_u)\big]}_{v_{\mathrm{shared}}} \nonumber\\
& +\;(1{+}s)\!\!\sum_{i\in S_c\setminus S_u}\!\! g_i^c E_i(h_c)\;-\;s\!\!\sum_{j\in S_u\setminus S_c}\!\! g_j^u E_j(h_u).
\label{eq:delta-decomp}
\end{align}
The exclusive unconditional term is a write that the conditional branch
did not produce in this pass. Let $\Vsc$ be the rank-$m$ section
of stacked conditional activations defined below, with projector $P_c$,
and define the leak
\begin{equation}
\delta(s) \;:=\; -s\,(I-P_c)\,v_u.
\label{eq:delta-id}
\end{equation}
Then $\vcfg = (1{+}s)v_c - s\,P_c v_u + \delta(s)$, and
$\lVert\delta(s)\rVert = s\,\lVert(I-P_c)v_u\rVert$.
The leak is exactly linear in $s$ and vanishes if and only if
$v_u\in\Vsc$. Exclusive experts are the MoE-specific source of this
residual (\appref{app:theory}); a dense FFN has $\Droute\equiv 0$ and
no exclusive experts.

If $v_c,v_u\in\Vc$ for any subspace $\Vc$, then $\vcfg\in\Vc$: CFG is
closed. We need neither $S_c=S_u$ nor a shared global basis. As in
\cref{fig:main}, it is enough that $v_u$ lie in the conditional
subspace. SAGE therefore penalizes only the component of the
unconditional write outside $\Vsc$,
\begin{equation}
\Lsage(\theta) := \E\!\left[\big\lVert
(I-P_c)\,f_{\mathrm{MoE}}(h_u)\big\rVert_2^2\right].
\label{eq:sage-loss}
\end{equation}

In a minibatch of $B$ samples with $T$ tokens per sample, stack the layer outputs into
$F_c,F_u\in\R^{d_h\times n}$ with $n=BT$. Full-span QR of $F_c$ is
vacuous whenever $n\ge d_h$. We therefore draw a column subset $\Omega$
of size
\begin{equation}
m=\min\bigl(n,\lfloor d_h/\ksplit\rfloor\bigr),
\label{eq:m-split}
\end{equation}
where the integer $\ksplit$ is chosen so that $\Vsc$ remains a
proper subspace whenever $n$ is large. Stop-gradient $F_c$ and take a
thin QR of the section $F_{c,\Omega}=\hat{F}_c R$:
\begin{equation}
\Lsage
= \frac{1}{\lVert F_u\rVert_F^2}
\big\lVert F_u - \hat{F}_c\hat{F}_c^{\!\top} F_u\big\rVert_F^2.
\label{eq:sage-batch}
\end{equation}
The cost is $\mathcal{O}(d_h m^2+d_h m n)$ per layer, and the
total objective is
\begin{equation}
\mathcal{L}_{\mathrm{total}}(\theta)
= \Lfm(\theta) + \lambda\sum_\ell \Lsage^{(\ell)}(\theta),\qquad\lambda>0.
\label{eq:total-loss}
\end{equation}
The batch leak energy is then $s\,\varepsilon\,\lVert F_u\rVert_F$
(\suppcref{thm:stability}), where $\varepsilon$ is the training residual. SAGE needs a paired unconditional forward on the same $x_t$ in order to
form $(F_c,F_u)$. The flow-matching term still follows standard CFG
dropout: with probability $p_{\mathrm{drop}}$ one supervises $v_u$,
otherwise $v_c$. Inference is unchanged. The procedure is
\cref{alg:sage}.

\begin{algorithm}[t]
\caption{MoE-DiT Training with SAGE}
\label{alg:sage}
\begin{algorithmic}[1]
\REQUIRE Dataset $\mathcal{D}$, MoE-DiT $v_\theta$, SAGE weight $\lambda$,
  CFG dropout $p_{\mathrm{drop}}$, split factor $\ksplit$
\FOR{each training iteration}
  \STATE Sample minibatch $\{(x_1, c)\}\sim\mathcal{D}$, $t\sim\mathcal{U}[0,1]$, $\varepsilon\sim\mathcal{N}(0,I)$
  \STATE $x_t\leftarrow t\,x_1+(1-t)\,\varepsilon$;\quad $y\leftarrow x_1-\varepsilon$
  \STATE Conditional pass: $v_c,\{f_{\mathrm{MoE}}^{(\ell)}(h_c)\}_\ell\leftarrow\mathrm{Forward}(v_\theta, x_t, t, c)$
  \STATE Unconditional pass: $v_u,\{f_{\mathrm{MoE}}^{(\ell)}(h_u)\}_\ell\leftarrow\mathrm{Forward}(v_\theta, x_t, t, \varnothing)$
  \STATE Draw $u\sim\mathrm{Bernoulli}(p_{\mathrm{drop}})$;\;
    $\Lfm\leftarrow\|v_u-y\|^2$ if $u$ else $\|v_c-y\|^2$
  \FOR{each MoE layer $\ell$}
    \STATE $F_c\leftarrow\mathrm{stack}(f_{\mathrm{MoE}}^{(\ell)}(h_c))$;\quad
      $F_u\leftarrow\mathrm{stack}(f_{\mathrm{MoE}}^{(\ell)}(h_u))$
      \COMMENT{$F_c,F_u\in\R^{d_h\times n}$, $n=BT$}
    \STATE $m\leftarrow\min(n,\lfloor d_h/\ksplit\rfloor)$;\;
      sample $m$ columns $\Omega$;\;
      $\hat F_c,\_\leftarrow\mathrm{thinQR}(\mathrm{sg}(F_{c,\Omega}))$
    \STATE $\Lsage^{(\ell)}\leftarrow\frac{1}{\|F_u\|_F^2}\,\|F_u-\hat F_c\hat F_c^{\!\top} F_u\|_F^2$
  \ENDFOR
  \STATE $\mathcal{L}_{\mathrm{total}}\leftarrow\Lfm+\lambda\sum_\ell\Lsage^{(\ell)}$
  \STATE $\theta\leftarrow\theta-\mathrm{lr}\cdot\nabla_\theta\mathcal{L}_{\mathrm{total}}$
\ENDFOR
\end{algorithmic}
\end{algorithm}

%% ====================================================================
%% 5. EXPERIMENTS
%% ====================================================================
\section{Experiments}
\label{sec:experiments}

We validate SAGE in two complementary settings. First, we use a 2-D Gaussian mixture toy experiment (\cref{sec:exp-toy}) to empirically verify the leak arising from CFG in MoE-DiTs. In this setting, the leak $\|\delta(s)\|$ is directly measurable, and the true likelihood is available in closed form. Subsequently, we demonstrate the effectiveness of SAGE on a large-scale, 1B-parameter MoE-DiT for text-to-image generation (\cref{sec:exp-t2i}), showing that our method successfully scales to high-dimensional data.

\subsection{Toy Experiment}
\label{sec:exp-toy}

We construct a 2-D class-conditional Gaussian mixture with $K{=}8$ isotropic modes ($r{=}5$, $\sigma{=}0.3$; $80{,}000$ training samples).
The backbone is a flow-matching~\citep{lipman2022fm} multilayer perceptron (MLP) with a single MoE layer ($N{=}8$ experts, top-$k{=}2$, $d_h{=}64$, $d_f{=}16$).
A dense FFN of width $32$ (matching the number of active parameters) serves as a control.
With $T{=}1$, SAGE uses every available conditional write, so $\ksplit$ is not needed.
We compare three configurations: Dense, MoE, and MoE + SAGE ($\lambda{=}1$), all trained with AdamW~\citep{loshchilov2019adamw} for $20{,}000$ steps and sampled using a $50$-step Euler ODE solver.

In this toy setup, conditional generation corresponds to class-specific point clouds, while the unconditional counterpart spans the remaining empty regions. Thus, applying CFG essentially concentrates the points of each class toward their respective centers. In theory, a larger guidance scale should drive the empirical cluster centers closer to the ground-truth centers. However, due to the inherent CFG bias, an excessively large guidance scale often causes the generated clusters to deviate from the ground-truth centers.

\begin{figure}[t]
\centering
\includegraphics[width=\linewidth]{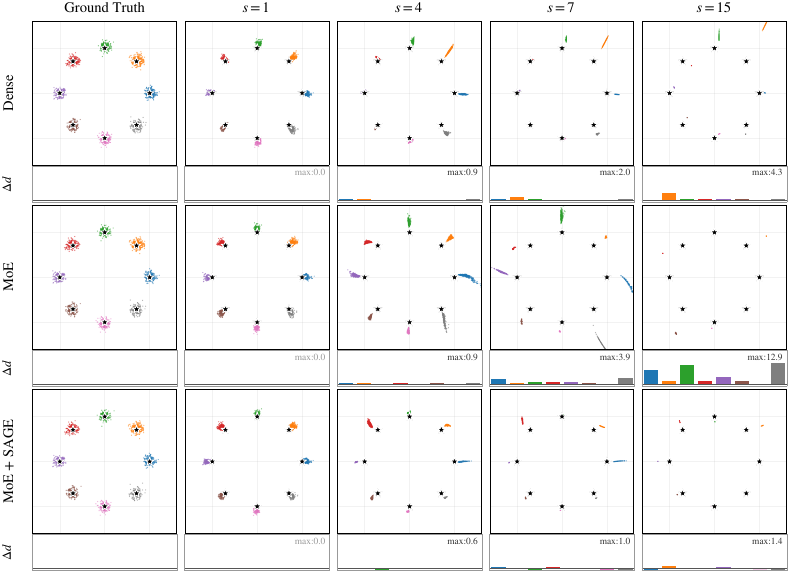}
\caption{\textbf{Toy experiment sample grids} for Dense, MoE, and MoE+SAGE across $s\in\{1,4,7,15\}$, with per-class center drift ($\Delta d$; lower is better). The baseline MoE exhibits increasing drift at large $s$; SAGE keeps the drift substantially smaller.}
\label{fig:samples}
\vspace{-0.4cm}
\end{figure}

\Cref{fig:samples} visualizes inference results for each model across four CFG scales, with per-class excess drift $\Delta d$ relative to $s{=}1$ shown as bar charts below each panel; the annotation in each bar chart reports the maximum $\Delta d$.
\Cref{tab:purity} quantifies two complementary metrics: the mean $\Delta d$ averaged over all eight classes (with the per-class maximum in parentheses, matching the bar-chart annotations in \cref{fig:samples}) and mode purity.
At $s{=}7$, the maximum per-class excess drift for MoE reaches $3.9$, over $1.9{\times}$ that of Dense, while MoE\,+\,SAGE stays at only $1.0$.
At $s{=}15$, the gap widens dramatically: the maximum per-class excess drift for MoE reaches $12.9$, while MoE\,+\,SAGE remains at $1.4$ (${9.2{\times}}$ down).
Mode purity tells the same story: MoE collapses to $0.121$ at $s{=}7$ while MoE\,+\,SAGE maintains $0.781$ (a ${6.5{\times}}$ improvement).
These results confirm that the subspace leak from routing misalignment is MoE-specific,
and that SAGE effectively neutralizes it.

\begin{table}[h]
\centering
\caption{\textbf{Quantitative results on the toy experiment.}
Excess drift $\Delta d$ reports the mean per-class distance increase relative to $s{=}1$ (matching the bar charts in \cref{fig:samples}); the per-class maximum is shown in parentheses for cross-reference with the bar-chart annotations.
Mode purity (higher is better) measures separation across the eight modes.
MoE drift explodes at high guidance while MoE\,+\,SAGE stays close to Dense.}
\label{tab:purity}
\small
\setlength{\tabcolsep}{10.5pt}
\begin{tabular}{@{}l ccc ccccc@{}}
\toprule
& \multicolumn{3}{c}{Excess Drift $\Delta d\;(\downarrow)$} & \multicolumn{5}{c}{Mode Purity $(\uparrow)$} \\
\cmidrule(lr){2-4} \cmidrule(lr){5-9}
Method & $s{=}4$ & $s{=}7$ & $s{=}15$
       & $s{=}1$ & $s{=}2$ & $s{=}4$ & $s{=}7$ & $s{=}15$ \\
\midrule
Dense       & 0.33\,{\scriptsize(0.9)} & 0.50\,{\scriptsize(2.0)} & 1.06\,{\scriptsize(4.3)}
            & .993 & .968 & .794 & .586 & .501 \\
MoE         & 0.72\,{\scriptsize(0.9)} & 1.83\,{\scriptsize(3.9)} & 5.59\,{\scriptsize(12.9)}
            & .989 & .929 & .394 & .121 & .125 \\
MoE\,+\,SAGE & 0.34\,{\scriptsize(0.6)} & 0.39\,{\scriptsize(1.0)} & 0.39\,{\scriptsize(1.4)}
            & .997 & .989 & .917 & .781 & .666 \\
\bottomrule
\end{tabular}
\end{table}

\subsection{Large-Scale Text-to-Image Generation}
\label{sec:exp-t2i}

\begin{table}[t]
\centering
\caption{\textbf{Overall scores on DPG-Bench and GenEval across CFG scales.}
SAGE consistently outperforms the Baseline at all $s \ge 3$ and degrades more
gracefully under high guidance, while the KL routing
constraint hurts performance across the board. $\Delta$\,(peak$\to$10) measures the change from the peak score to the score at $s{=}10$; $\Delta_{\mathrm{base}}$ shows the difference from the Baseline's peak score. \textbf{Bold} denotes the best score.}
\label{tab:cfg-scales}
\small
\setlength{\tabcolsep}{8.9pt}
\renewcommand{\arraystretch}{1.15}
\begin{tabular}{@{}ll ccccc @{\hspace{8pt}} c @{\hspace{8pt}} c@{}}
\toprule
 & & \multicolumn{5}{c}{CFG guidance scale $s$} & & \\
\cmidrule(lr){3-7}
Benchmark & Method & 1 & 3 & 5 & 7 & 10 & $\Delta$\,(peak$\to$10) & $\Delta_{\mathrm{base}}$ \\
\midrule
\multirow{3}{*}{DPG-Bench}
 & Baseline                & \textbf{54.77} & 61.76 & 56.33 & 52.38 & 48.56 & $-13.20$ & -- \\
 & SAGE ($\lambda{=}1$)    & 52.30 & \textbf{67.49} & \textbf{65.63} & \textbf{63.36} & \textbf{59.55} & $-7.94$ & $+5.73$ \\
 & KL ($\lambda_{\mathrm{kl}}{=}1$) & 40.95 & 50.65 & 46.82 & 44.38 & 41.08 & $-9.57$ & $-11.11$ \\
\midrule
\multirow{3}{*}{GenEval}
 & Baseline                & \textbf{30.3} & 57.1 & 55.5 & 53.2 & 47.1 & $-10.0$ & -- \\
 & SAGE ($\lambda{=}1$)    & 26.5 & \textbf{59.8} & \textbf{59.3} & \textbf{57.1} & \textbf{52.5} & $-7.3$ & $+2.7$ \\
 & KL ($\lambda_{\mathrm{kl}}{=}1$) & 13.8 & 36.0 & 34.5 & 32.2 & 29.5 & $-6.5$ & $-21.1$ \\
\bottomrule
\end{tabular}
\end{table}

\begin{figure}[t]
\centering
\includegraphics[width=1.0\linewidth]{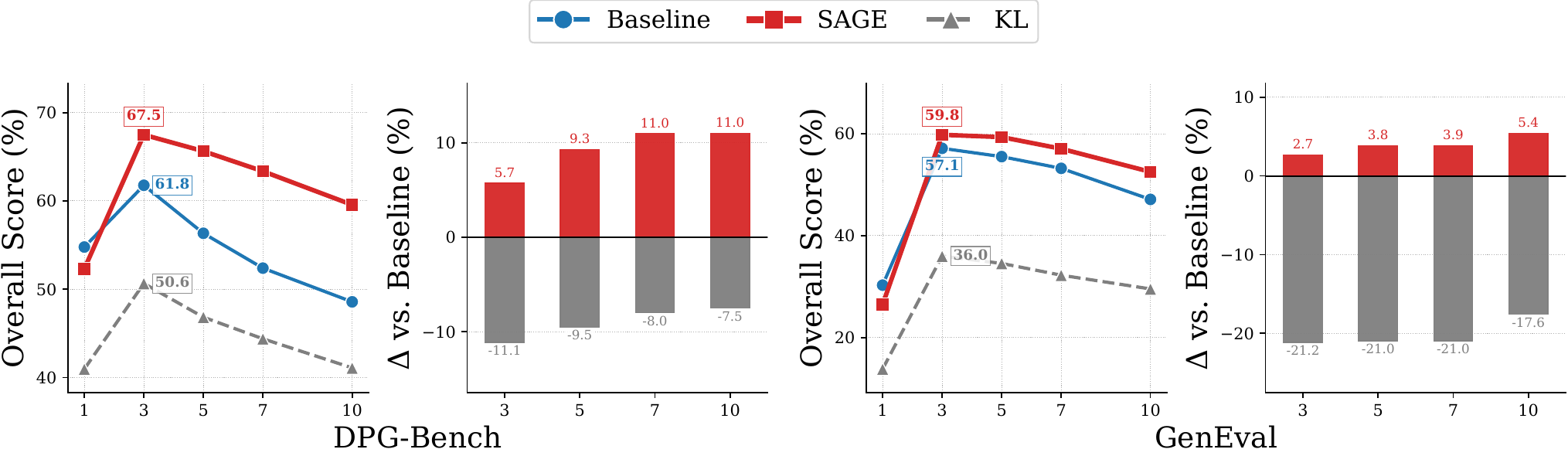}
\caption{\textbf{Performance improvements of SAGE across CFG scales.} SAGE (red) maintains high performance across the full guidance range, while the Baseline (blue) peaks at $s{=}3$ and declines sharply. The KL routing constraint (gray) consistently underperforms both methods, confirming that naive distributional alignment of routing decisions is counterproductive.}
\label{fig:cfg-curves}
\vspace{-0.2cm}
\end{figure}

\subsubsection{Setup}
\label{sec:exp-t2i-setup}

We train a 1B-parameter MoE-DiT consisting of 12 Transformer layers~\citep{vaswani2017attention} with hidden dimension $d_h{=}1536$ and 12 attention heads (128 dimensions per head).
Each layer contains 8 routed experts and 1 shared expert, with top-2 routing applied to the routed experts. The intermediate dimensions of the routed and shared experts are 1024 and 2048, respectively. SAGE is applied with $\ksplit{=}2$.
Text conditioning is provided by a frozen Qwen2.5-VL-7B-Instruct encoder~\citep{bai2025qwen25vl}.

We curate a 100M-image subset of LAION-5B~\citep{schuhmann2022laion5b}, re-caption it with Qwen2.5-VL 72B, and resize and crop all images to a resolution of approximately $256 \times 256$. We use AdamW with a learning rate of $10^{-4}$, a constant learning-rate schedule following a $1{,}000$-step warmup, and a global batch size of 512 on 32 NVIDIA H200 GPUs.
We set $p_{\mathrm{drop}}{=}0.1$, the MoE auxiliary load-balancing loss coefficient~\citep{fedus2021switch} to $\alpha_{\mathrm{aux}}{=}0.001$, and the router z-loss coefficient~\citep{zoph2022stmoe} to $\alpha_z{=}0.001$.
Three configurations are compared: Baseline ($\lambda{=}0$), SAGE ($\lambda{=}1$), and KL Routing Constraint ($\lambda_{\mathrm{kl}}{=}1$), which penalizes the KL divergence between the conditional and unconditional routing distributions to force $S_c \approx S_u$.
All models are trained for the same 200,000 steps to ensure a fair comparison.

We use two benchmarks: (a)~DPG-Bench~\citep{dpgbench2024}, evaluated with an mPLUG backbone~\citep{li2022mplug}, for which we report the overall score and a per-category breakdown; and (b)~GenEval~\citep{ghosh2024geneval}, with 553 prompts and 4 seeds per prompt.
CFG scales are swept over $s \in \{1.0,\,3.0,\,5.0,\,7.0,\,10.0\}$.

\subsubsection{Main Results}
\label{sec:exp-t2i-main}

\begin{figure}[t]
\centering
\includegraphics[width=0.49\linewidth]{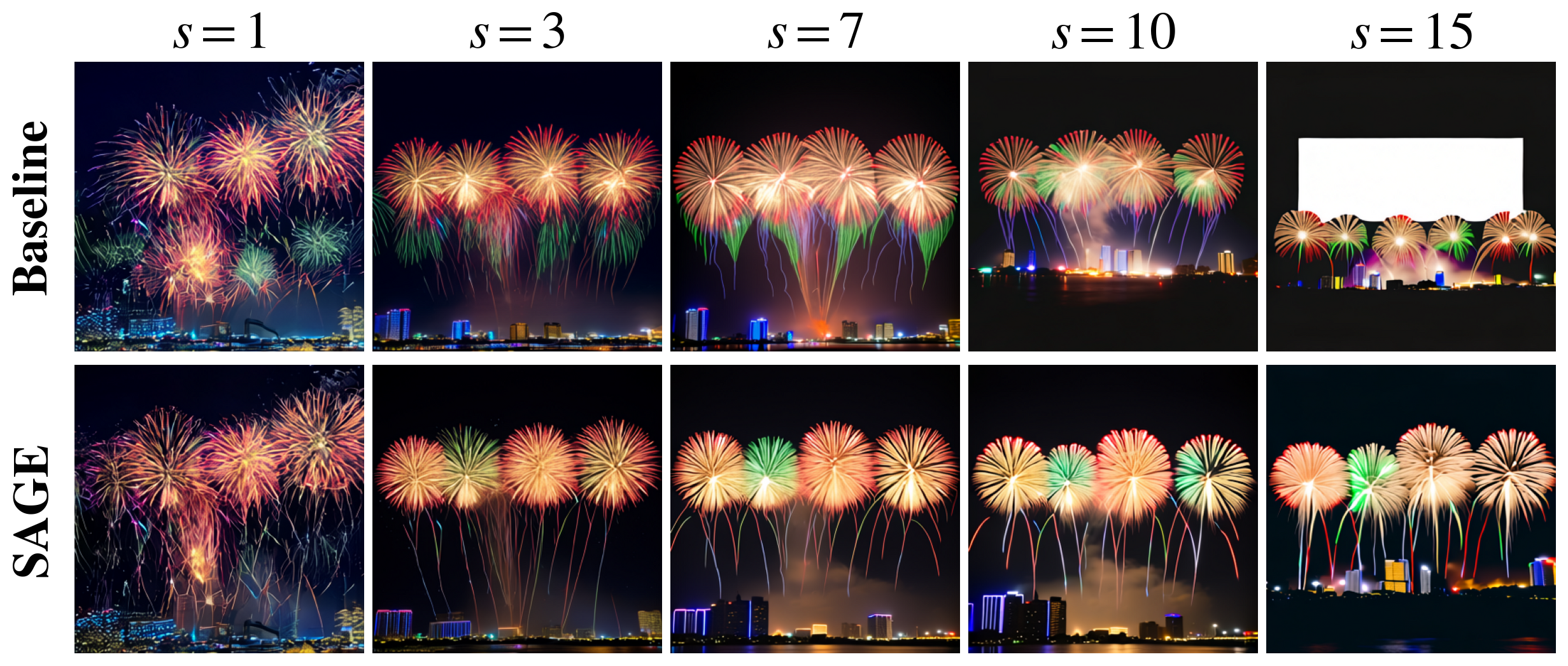}\hfill
\includegraphics[width=0.49\linewidth]{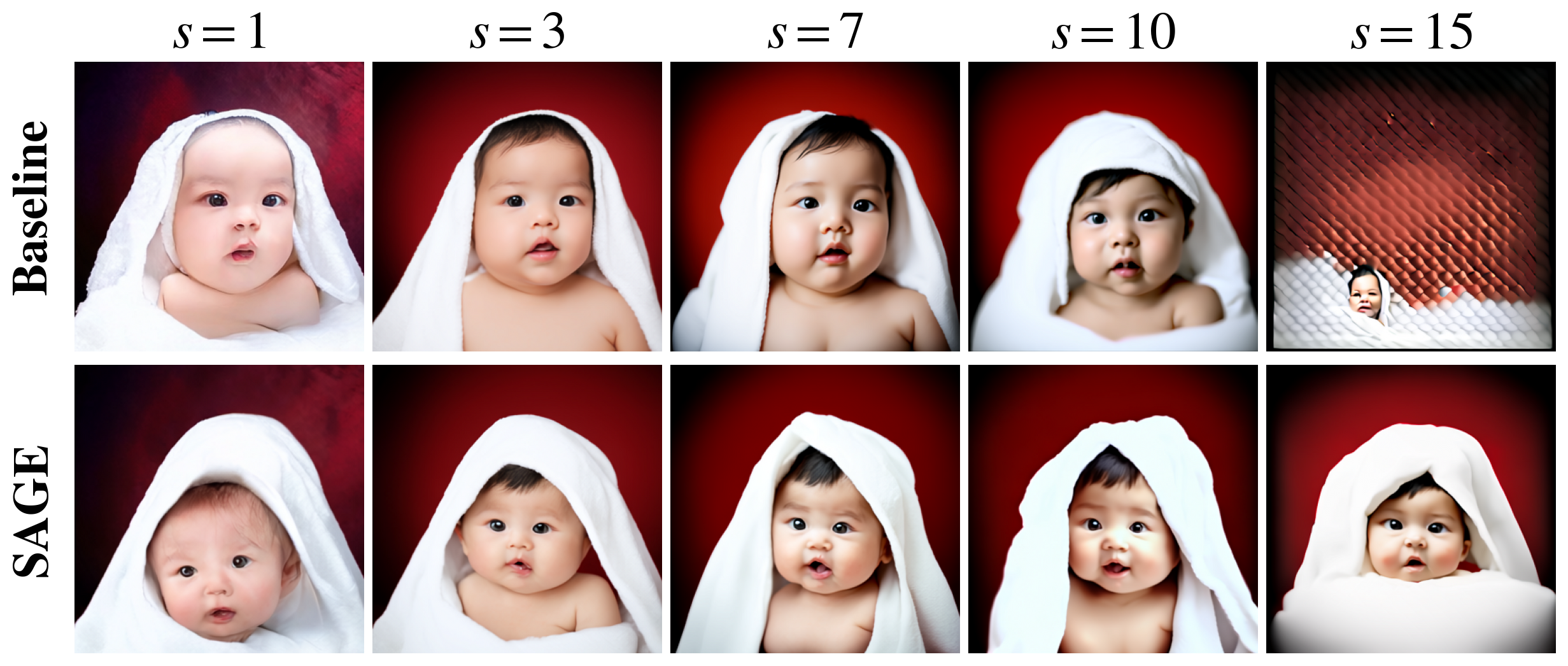}\\[2pt]
\includegraphics[width=0.49\linewidth]{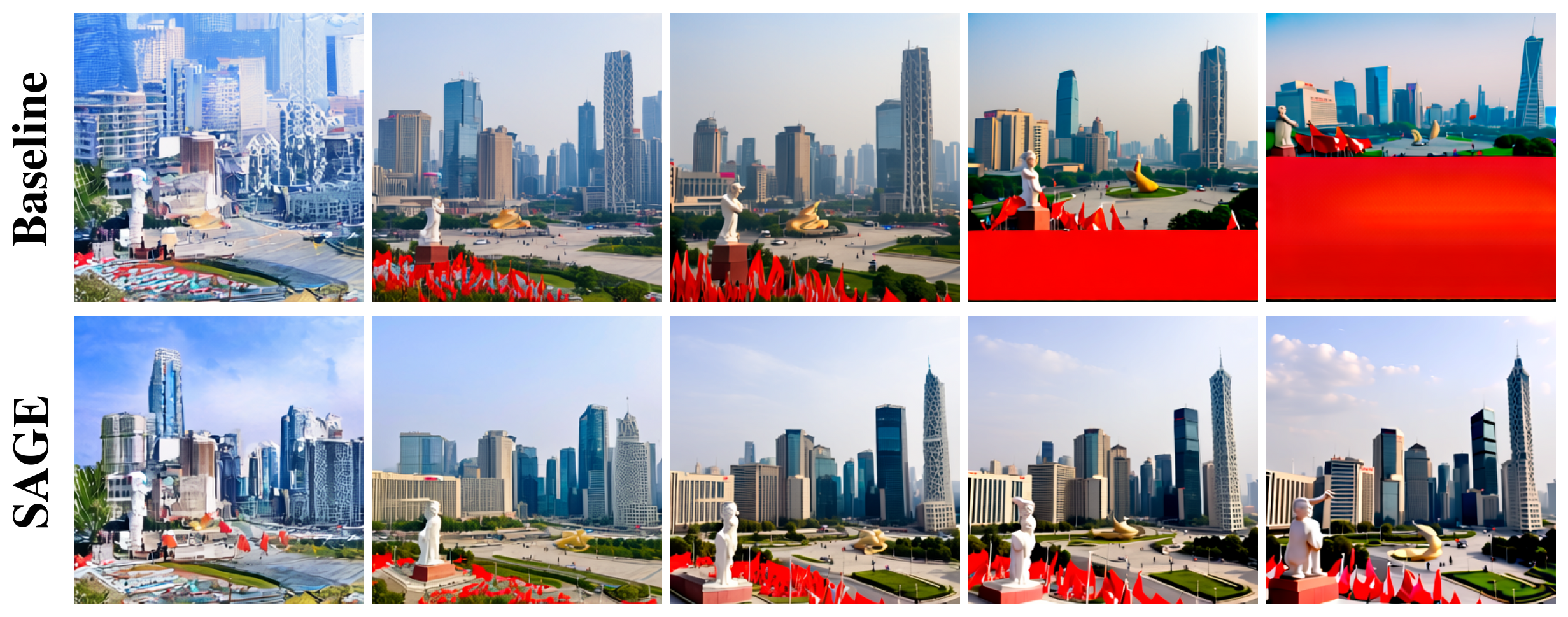}\hfill
\includegraphics[width=0.49\linewidth]{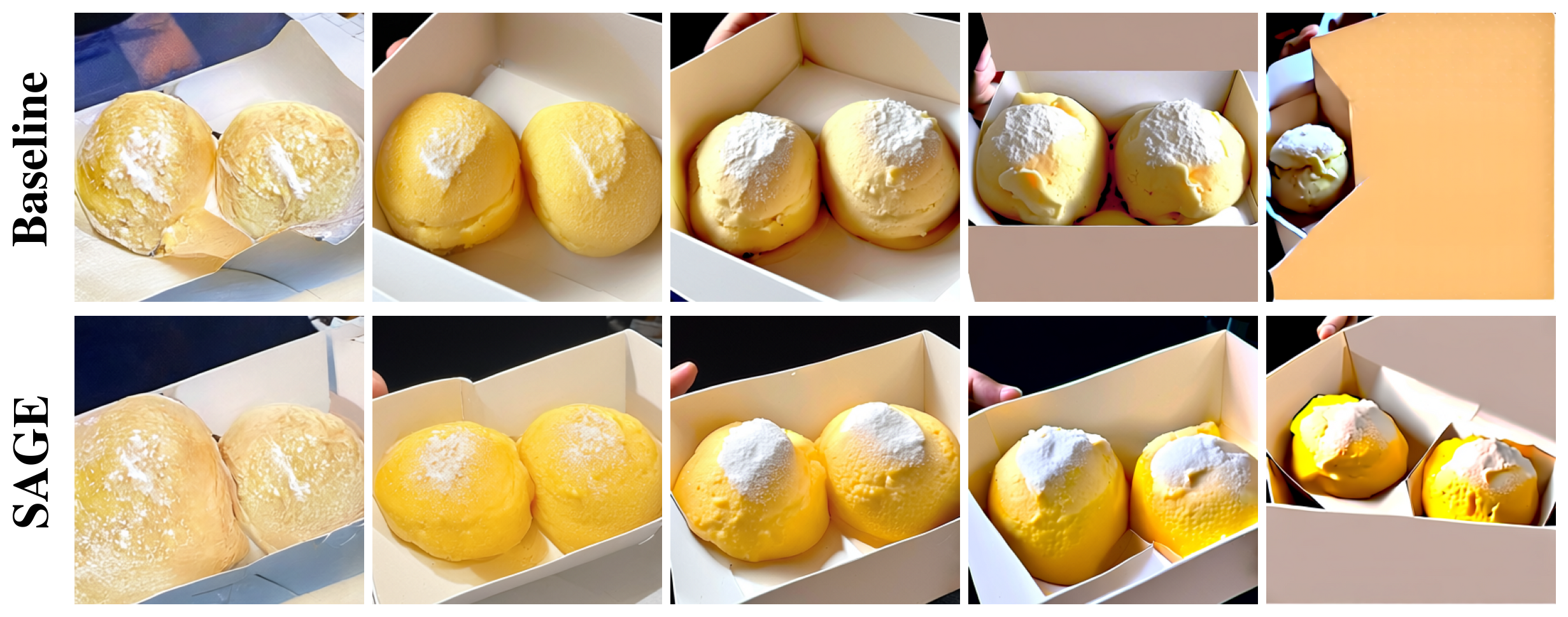}\\[2pt]
\includegraphics[width=0.49\linewidth]{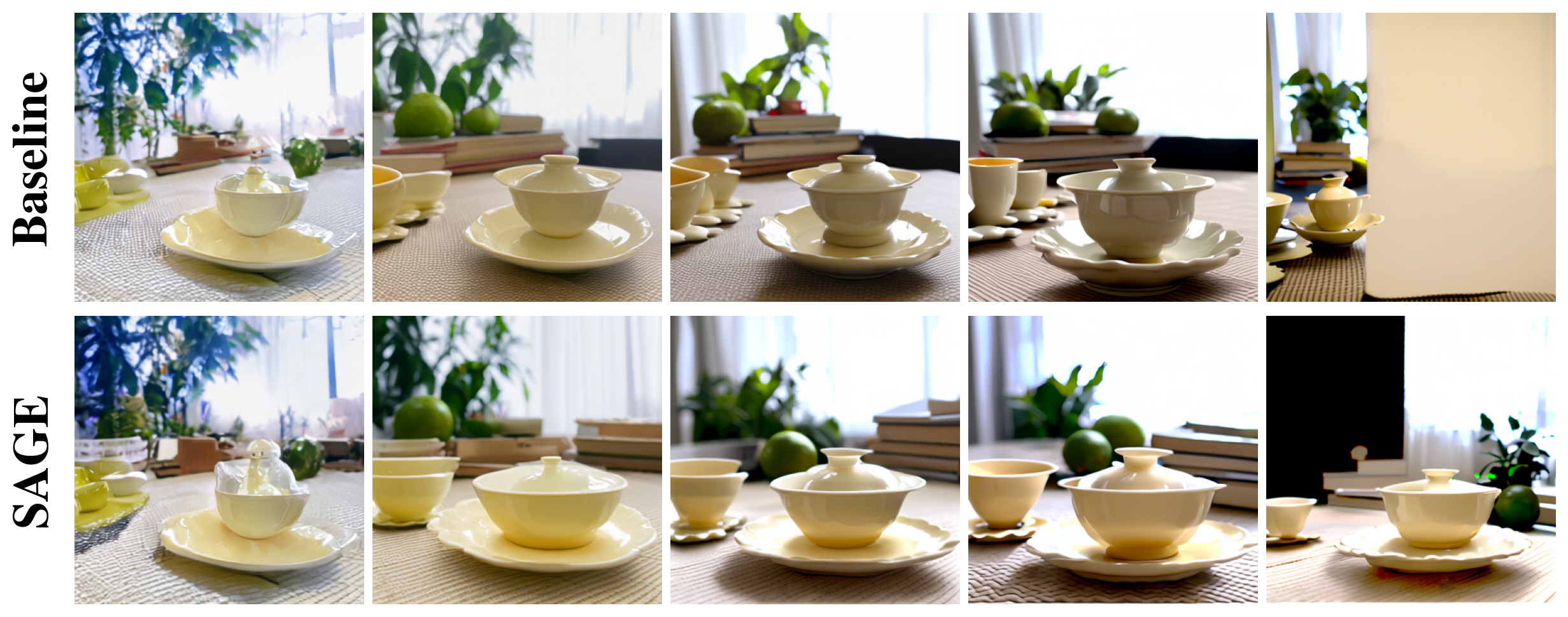}\hfill
\includegraphics[width=0.49\linewidth]{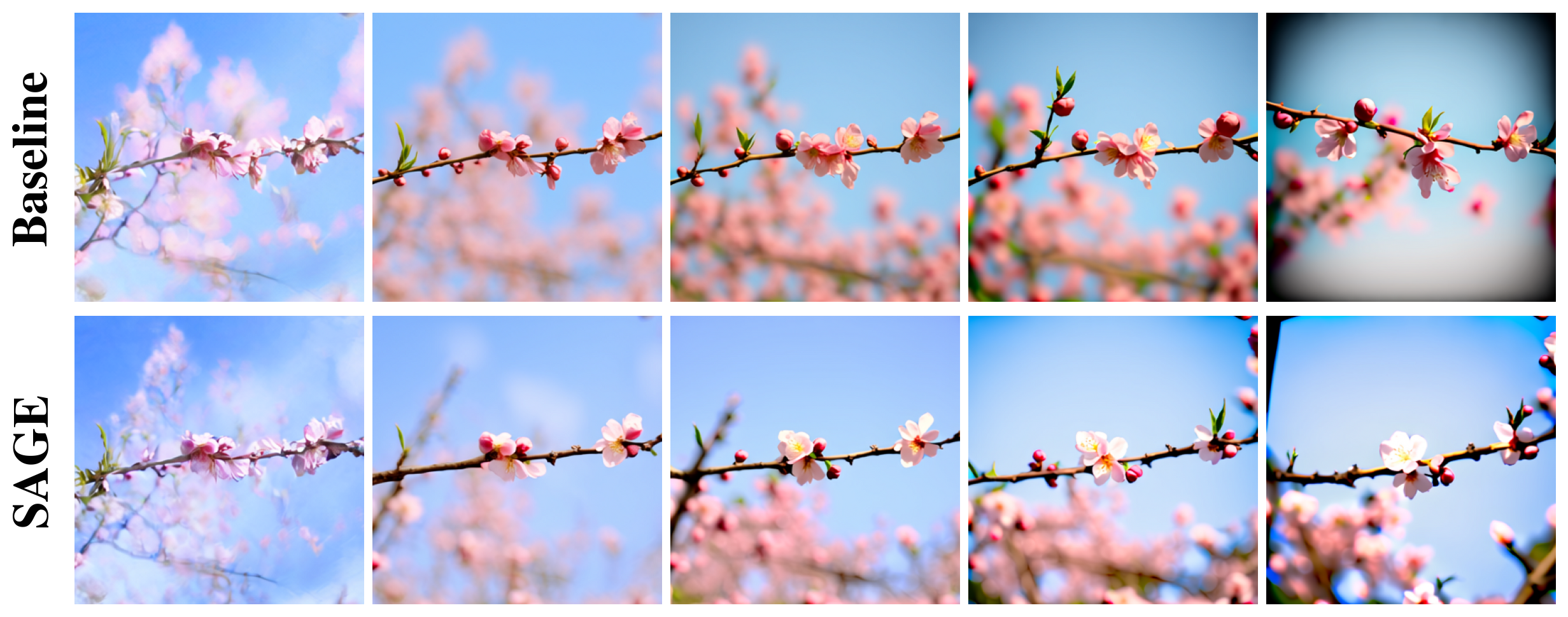}
\caption{\textbf{Qualitative comparison across CFG scales.} Each panel shows the same prompt rendered by Baseline (top) and SAGE (bottom) at $s \in \{1, 3, 7, 10, 15\}$ from left to right. At $s{=}10$ and $15$, the Baseline exhibits over-saturation, white patches, and structural collapse, while SAGE preserves coherent composition and realistic textures.}
\label{fig:qualitative-t2i}
\end{figure}

\paragraph{Quantitative comparison.}

The ordering SAGE $>$ Baseline $>$ KL holds consistently at all $s \ge 3$ on both benchmarks (\cref{tab:cfg-scales,fig:cfg-curves}). SAGE significantly enhances generation quality, boosting peak DPG-Bench performance by 9.3\%. Meanwhile, SAGE's robustness is particularly notable: the DPG-Bench score drops by only $7.94$ points from its peak to $s{=}10$, versus $13.20$ for the Baseline; moreover, SAGE at $s{=}7$ still exceeds the Baseline's peak.
The DPG-Bench evaluation protocol and category-level breakdown are provided in \appref{app:dpg-detail}.
On GenEval, the pattern is the same: SAGE retains a $+5.4$\,pp advantage at $s{=}10$ (52.5\% vs.\ 47.1\%) and loses only $7.3$\,pp from its peak, compared with $10.0$\,pp for the Baseline.
The largest per-task improvement is on color-attribute binding ($+11.5$\,pp; full breakdown in \appref{app:geneval-detail}), a compositional task that relies heavily on precise conditional control, exactly where a write outside the conditional subspace is most damaging. To confirm that the CFG-robustness benefit of SAGE also holds for distributional image fidelity, we report FID~\citep{heusel2017fid} across CFG scales in \appref{app:fid}.

The KL routing constraint, which penalizes $D_{\mathrm{KL}}(\pi_c \| \pi_u)$ to force consistent routing, is counterproductive on every metric.
By collapsing the routing distributions, it destroys expert specialization: the model can no longer assign different experts to different semantic attributes, resulting in a $-21$\,pp drop on GenEval relative to the Baseline at $s{=}3$.
This provides strong empirical evidence for the theoretical insight of \cref{sec:method-sage}: the correct fix is subspace alignment, not routing agreement.
In contrast to the KL constraint, SAGE leaves the router essentially untouched: its conditional and unconditional branches stay as routing-diverse as the Baseline at every guidance scale, so subspace alignment is achieved without sacrificing expert specialization (\appref{app:routing}).

\paragraph{Qualitative comparison.}
\Cref{fig:qualitative-t2i} presents several prompts across the full CFG sweep.
At moderate guidance ($s \le 7$), both methods produce plausible images; as $s$ increases to $10$ and $15$, the Baseline develops over-saturated colors, white-out patches, and structural repetition, all symptomatic of the subspace leak $\delta(s)$ in~\eqref{eq:delta-id}.
SAGE preserves coherent composition, fine-grained texture, and realistic color balance even at $s{=}15$.
The quantitative and qualitative trends therefore agree: the benefit of SAGE is not a benchmark artifact but a visible improvement in image structure at high guidance.

\subsubsection{Ablation Study}
\label{sec:exp-ablation}

Keeping the setup of \cref{sec:exp-t2i-setup} fixed and evaluating at guidance scales $s\in\{3,5\}$, we ablate the two most consequential design choices of SAGE: \emph{when} the SAGE loss is applied and \emph{how strongly} it is weighted.
For both ablations, each configuration is evaluated on DPG-Bench and GenEval at each scale.
Two further ablations, covering \emph{how often} the loss needs to be activated and \emph{how much} of its effect a shared expert already provides for free, are deferred to \appref{app:ablations} and use the same two guidance scales.

\paragraph{Training timing.}
A natural question is whether SAGE can be applied post-hoc to an already-trained baseline by fine-tuning with $\Lsage$ for a small number of additional steps. \Cref{tab:posthoc_lambda} compares SAGE trained from scratch with post-hoc fine-tuning for 5k and 10k steps, all with $\lambda{=}1$. Applying SAGE from the start yields better performance, whereas post-hoc fine-tuning fails to improve on the baseline.
We hypothesize that once the two branches' activation clouds have specialized apart during pre-training, a short SAGE fine-tune cannot pull $F_u$ into the conditional section without harming the flow-matching fit: the residual $\range(F_u)\not\subseteq\Vsc$ has already crystallized, and rotating the writes into containment requires training from scratch.

\paragraph{Loss weight sensitivity.}
To assess sensitivity to the loss weight at scale, we vary $\lambda \in \{0.5, 1, 2.0, 5.0\}$. \cref{tab:posthoc_lambda} shows that the default weight $\lambda{=}1$ performs best on both benchmarks at both guidance scales. Relative to the baseline in the left panel, reducing the weight to $0.5$ yields only small gains, while increasing it to $2.0$ or $5.0$ lowers all four scores below the baseline. This peaked response is consistent with a trade-off between subspace alignment and flow-matching fit: an excessively large SAGE loss weight may restrict the representations needed for accurate velocity prediction. \suppcref{thm:stability} identifies leak energy with the SAGE residual; it does not imply that benchmark scores improve monotonically with $\lambda$.

\paragraph{Other ablations.}
\appref{app:ablations} reports further ablations on activation frequency and the shared expert, together with a comparison against ProMoE~\citep{promoe2025}, at $s\in\{3,5\}$.
Activating SAGE every second or fifth training step retains most of the gains from applying it every step, indicating that continuous activation is unnecessary. Removing the shared expert lowers baseline performance, but SAGE still improves both benchmarks, supporting the view that a shared expert alone does not replace explicit subspace alignment.
ProMoE improves over the baseline but remains below the default SAGE configuration at both guidance scales.

\begin{table}[t]
\centering
\caption{\textbf{Training timing and loss weight analysis.}
\textbf{Left:} comparison of from-scratch SAGE and post-hoc fine-tuning of a pre-trained baseline for 5k or 10k steps, with $\lambda{=}1$ for all SAGE variants.
\textbf{Right:} effect of the SAGE loss weight $\lambda$; the default $\lambda{=}1$ performs best among the tested values on both benchmarks at both guidance scales.}
\label{tab:posthoc_lambda}
\small
\setlength{\tabcolsep}{3pt}
\begin{minipage}[t]{0.56\linewidth}
\centering
\begin{tabular}{@{}lcccc@{}}
\toprule
& \multicolumn{2}{c}{$s{=}3$} & \multicolumn{2}{c}{$s{=}5$} \\
\cmidrule(lr){2-3} \cmidrule(lr){4-5}
Method & DPG & GenEval & DPG & GenEval \\
\midrule
Baseline & 61.76 & 57.1 & 56.33 & 55.5 \\
SAGE (from scratch) & 67.49 & 59.8 & 65.63 & 59.3 \\
Post-hoc SAGE (5k) & 61.32 & 56.3 & 56.23 & 54.8 \\
Post-hoc SAGE (10k) & 60.20 & 55.3 & 55.73 & 53.5 \\
\bottomrule
\end{tabular}
\end{minipage}\hfill
\begin{minipage}[t]{0.41\linewidth}
\centering
\begin{tabular}{@{}lcccc@{}}
\toprule
& \multicolumn{2}{c}{$s{=}3$} & \multicolumn{2}{c}{$s{=}5$} \\
\cmidrule(lr){2-3} \cmidrule(lr){4-5}
$\lambda$ & DPG & GenEval & DPG & GenEval \\
\midrule
0.5 & 65.31 & 58.8 & 62.07 & 57.8 \\
1 & {67.49} & {59.8} & {65.63} & {59.3} \\
2.0 & 64.85 & 58.6 & 63.33 & 58.2 \\
5.0 & 59.94 & 56.2 & 58.69 & 55.8 \\
\bottomrule
\end{tabular}
\end{minipage}
\end{table}

%% ====================================================================
%% 6. CONCLUSION
%% ====================================================================
\section{Conclusion and Future Work}
\label{sec:conclusion}

We identified a subspace leak in MoE diffusion models. Routing misalignment lets the two CFG branches realize different subspaces; exclusive experts let the unconditional write leave the conditional subspace, and CFG amplifies that residual linearly in $s$.
We proposed SAGE, which penalizes the component of the unconditional MoE write outside $\Vsc$ at training time, requires no architectural change, and incurs zero inference cost. From a controlled toy microscope to a 1B text-to-image model, SAGE consistently improves CFG stability.

Our large-scale evaluation covers models up to the 1B-parameter scale. Scaling SAGE to 100B+ models, analyzing how errors compound across layers in deeper architectures, and exploring video generation remain important directions for future work. Furthermore, CFG mechanisms in most generative models are considerably more complex; for instance, text-to-audio-video (T2AV) and reference-to-video (R2V) tasks often employ double CFG and offer promising settings for future investigation.

\newpage

\bibliography{iclr2027_conference}
\bibliographystyle{iclr2027_conference}
\clearpage
\appendix
\section{Theoretical Analysis}
\label{app:theory}

This section records the identities used in \maincref{sec:method}. CFG would mix the layer
writes $v_c=f_{\mathrm{MoE}}(h_c)$ and $v_u=f_{\mathrm{MoE}}(h_u)$. Notation follows
the main text: $S_c=\supp\,g(h_c)$, $S_u=\supp\,g(h_u)$,
$\Droute=1-|S_c\cap S_u|/k$, and
\begin{equation}
E_i(h)=W_2^{(i)}\sigma(W_1^{(i)}h+b_1^{(i)})+b_2^{(i)},
\quad W_2^{(i)}\in\R^{d_h\times d_f}.
\label{eq:expert}
\end{equation}
When a shared expert is present it is included in $f_{\mathrm{MoE}}$
and in the stacked activations below.

\subsection{Independent routing, different subspaces}

The two CFG branches route independently, so $S_c$ and $S_u$ need not
coincide. Stack the minibatch writes into $F_c,F_u\in\R^{d_h\times n}$,
$n=BT$. Let $\Omega$ be a set of
$m=\min(n,\lfloor d_h/\ksplit\rfloor)$ column indices
(\maineqref{eq:m-split}), and define the
conditional activation subspace
\begin{equation}
\Vsc \;:=\; \mathrm{span}(F_{c,\Omega}),
\qquad
P_c \;:=\; \hat F_c\hat F_c^{\!\top},
\label{eq:Ac-def}
\end{equation}
where $\hat F_c$ is an orthonormal basis of $\Vsc$ from thin QR of
$F_{c,\Omega}$ (stop-gradient on $F_c$, as in \maineqref{eq:sage-batch}).
Then $\dim\Vsc\le m$, so $\Vsc$ is a proper subspace whenever
$m<d_h$. It is \emph{not} claimed to contain every column of $F_c$
when $n>m$. The object SAGE uses is this realized subspace $\Vsc$.

\begin{lemma}[CFG closure]
\label{lem:closure}
If $v_c,v_u\in\Ac$ for any subspace $\Ac\subseteq\R^{d_h}$, then
$\vcfg=(1{+}s)v_c-s v_u\in\Ac$.
\end{lemma}

\begin{definition}[Subspace leak]
\label{def:leak}
Let $\videal:=(1{+}s)v_c-s\,P_c v_u$. The leak is the difference
\begin{equation}
\delta(s)
\;:=\;
\vcfg-\videal
\;=\;
-s\,(I-P_c)\,v_u.
\label{eq:delta-def}
\end{equation}
\end{definition}

Equivalently,
\begin{equation}
\vcfg
=(1{+}s)v_c-s\,P_c v_u+\delta(s),
\qquad
\lVert\delta(s)\rVert
=s\,\lVert(I-P_c)v_u\rVert.
\label{eq:leak-norm}
\end{equation}
So $\delta(s)=0$ if and only if $v_u\in\Vsc$, and the leak, when
present, is exactly linear in $s$. Independent routing makes
$v_u\notin\Vsc$ typical: the unconditional branch can write in
directions the conditional activations did not realize. SAGE drives
$\delta\to 0$; it does not claim $\vcfg\in\Vsc$ when $n>m$, because
tokens of $F_c$ outside $\Omega$ need not lie in $\Vsc$. Closure
(\cref{lem:closure}) is the motivation for aligning $v_u$ to a realized
conditional subspace, not a claim that the implemented section contains
the whole batch.

\subsection{What SAGE minimizes}

The implemented loss \maineqref{eq:sage-batch} is the relative energy of
$F_u$ outside $\Vsc$. Writing $\Delta$ for the matrix whose
columns are the per-token leaks \eqref{eq:delta-def}, the identity
\begin{equation}
\Delta
=-s\,(I-P_c)F_u,
\qquad
\lVert\Delta\rVert_F
=s\,\lVert(I-P_c)F_u\rVert_F
\label{eq:sage-frobenius}
\end{equation}
holds at every parameter value. With
\begin{equation}
\Lsage
=\frac{1}{\lVert F_u\rVert_F^2}
\lVert(I-P_c)F_u\rVert_F^2,
\label{eq:sage-rel}
\end{equation}
one has $\lVert\Delta\rVert_F=s\sqrt{\Lsage}\,\lVert F_u\rVert_F$.

\begin{proposition}[SAGE controls leak energy]
\label{thm:stability}
At any parameter value, $\lVert\Delta\rVert_F=s\,\lVert(I-P_c)F_u\rVert_F$.
In particular, if $\Lsage(\theta^*)=\varepsilon^2$, then
\begin{equation}
\lVert\Delta\rVert_F
=s\,\varepsilon\,\lVert F_u\rVert_F.
\label{eq:delta-sage}
\end{equation}
\end{proposition}

No Markov bound and no $\lambda^{-1/2}$ rate are claimed:
$\lambda$ only trades $\Lsage$ against $\Lfm$ during optimization
(\maincref{sec:exp-ablation}). Equation~\eqref{eq:sage-frobenius} is
the identity that \maineqref{eq:sage-batch} drives to zero.

\subsection{Exclusive experts}

Split $v_u=v_u^{\mathrm{sh}}+v_u^{\mathrm{ex}}$ into experts in
$S_c\cap S_u$ and in $S_u\setminus S_c$:
\begin{equation}
v_u^{\mathrm{ex}}
=\sum_{j\in S_u\setminus S_c} g_j^u\,E_j(h_u).
\label{eq:ex-mass}
\end{equation}
This extra write is MoE-specific: a dense FFN has $S_c=S_u=\{1\}$,
hence $\Droute\equiv 0$ and $v_u^{\mathrm{ex}}=0$. The split does
\emph{not} bound $\lVert(I-P_c)v_u\rVert$. Shared-set writes
$E_i(h_u)$ need not lie in $\Vsc$ (which is spanned by $E_i(h_c)$
and other tokens), and exclusive writes need not be orthogonal to
$\Vsc$. Independent routing only makes a residual outside $\Vsc$ typical.
Dense CFG can still fail for other reasons; SAGE targets the component
of $v_u$ outside $\Vsc$.

\subsection{Shared expert}
\label{app:shared}

A shared expert $E_0$ is evaluated on both branches and does not enter
$\Droute$. It reduces exclusive routed mass in $v_u^{\mathrm{ex}}$,
but it does not put $v_u$ in $\Vsc$: $E_0(h_u)$ still differs from
$E_0(h_c)$, and exclusive routed writes remain. Our 1B implementation
applies \maineqref{eq:sage-batch} to the full MoE output (shared plus
routed). Both experiments minimize the same loss: the toy model has
$T=1$ and uses every available conditional write, so $\ksplit$ is not
needed; the 1B model has $n=BT\gg d_h$ and uses $\ksplit{=}2$, so only
the rank-$m$ section is a proper subspace.

%% ====================================================================
%% APPENDIX B: ADDITIONAL EXPERIMENTAL RESULTS
%% ====================================================================
\section{Additional Experimental Results}
\label{app:extra-exp}

\subsection{DPG-Bench Evaluation and Category Breakdown}
\label{app:dpg-detail}

We evaluate the Baseline, SAGE ($\lambda{=}1$), and the KL routing constraint ($\lambda_{\mathrm{kl}}{=}1$) on DPG-Bench~\citep{dpgbench2024} at CFG scales~\citep{ho2022cfg} $s\in\{1,3,5,7,10\}$, using mPLUG~\citep{li2022mplug} for visual question answering (VQA).
For each prompt, we generate four $320\times320$ images using 50 sampling steps. 

\Cref{tab:dpg-detail} reports the five top-level categories: Global, Entity, Attribute, Relation, and Other.
SAGE exceeds the Baseline in all five categories at every tested $s\ge3$.
At $s{=}3$, the largest gains are in Other ($+6.0$\,pp) and Entity ($+4.7$\,pp); at $s{=}10$, the gains reach $+10.8$\,pp for Entity and $+8.1$\,pp for Relation.
Thus, the advantage under stronger guidance extends across object presence, attributes, and relations, rather than being confined to one category.

\Cref{tab:dpg-fine} gives all 13 second-level categories at $s{=}3$ and $s{=}5$.
SAGE improves over the Baseline in every subcategory at both scales.
The gains in counting are $+6.4$ and $+12.4$\,pp, respectively, while whole-entity presence improves by $+5.1$ and $+8.6$\,pp.
Size, texture, and spatial relations also improve at both scales, showing that the benefits cover multiple aspects of compositional generation.

\begin{table}[ht]
\centering
\caption{\textbf{DPG-Bench category scores (\%) across CFG scales.} Category scores aggregate raw per-question answers over all four images per prompt. Overall uses dependency-aware scoring and matches \maincref{tab:cfg-scales}; it is not the category average. Higher is better.}
\label{tab:dpg-detail}
\small
\setlength{\tabcolsep}{4pt}
\begin{tabular}{@{}ll cccccc@{}}
\toprule
Method & $s$ & Global & Entity & Attribute & Relation & Other & Overall \\
\midrule
\multirow{5}{*}{Baseline}
 & 1  & 70.06 & 68.04 & 73.16 & 84.98 & 43.00 & 54.77 \\
 & 3  & 73.10 & 73.87 & 81.00 & 85.71 & 57.40 & 61.76 \\
 & 5  & 65.43 & 69.66 & 77.44 & 82.28 & 54.60 & 56.33 \\
 & 7  & 61.02 & 66.10 & 75.00 & 79.96 & 52.00 & 52.38 \\
 & 10 & 59.65 & 62.59 & 71.66 & 77.90 & 50.60 & 48.56 \\
\midrule
\multirow{5}{*}{SAGE ($\lambda{=}1$)}
 & 1  & 69.45 & 65.94 & 71.04 & 84.91 & 42.30 & 52.30 \\
 & 3  & 76.44 & 78.58 & 83.36 & 88.65 & 63.40 & 67.49 \\
 & 5  & 73.40 & 77.70 & 81.54 & 86.95 & 65.10 & 65.63 \\
 & 7  & 70.44 & 76.10 & 80.20 & 86.60 & 63.90 & 63.36 \\
 & 10 & 66.49 & 73.34 & 77.21 & 85.96 & 62.00 & 59.55 \\
\midrule
\multirow{5}{*}{KL ($\lambda_{\mathrm{kl}}{=}1$)}
 & 1  & 63.30 & 55.26 & 66.73 & 78.74 & 29.30 & 40.95 \\
 & 3  & 67.63 & 64.66 & 77.45 & 80.32 & 48.90 & 50.65 \\
 & 5  & 64.44 & 61.86 & 74.45 & 77.10 & 45.60 & 46.82 \\
 & 7  & 60.56 & 59.56 & 72.40 & 74.76 & 44.10 & 44.38 \\
 & 10 & 59.80 & 56.13 & 69.33 & 73.60 & 42.20 & 41.08 \\
\bottomrule
\end{tabular}
\end{table}

\begin{table}[ht]
\centering
\caption{\textbf{Fine-grained DPG-Bench scores (\%) at $s\in\{3,5\}$.} All 13 second-level categories use the same four-image, pre-dependency aggregation as \cref{tab:dpg-detail}. SAGE and KL use $\lambda{=}1$ and $\lambda_{\mathrm{kl}}{=}1$, respectively. Higher is better.}
\label{tab:dpg-fine}
\small
\setlength{\tabcolsep}{4pt}
\begin{tabular}{@{}ll ccc ccc@{}}
\toprule
& & \multicolumn{3}{c}{$s{=}3$} & \multicolumn{3}{c}{$s{=}5$} \\
\cmidrule(lr){3-5} \cmidrule(lr){6-8}
Category & Subcategory & Baseline & SAGE & KL & Baseline & SAGE & KL \\
\midrule
Global & -- & 73.10 & 76.44 & 67.63 & 65.43 & 73.40 & 64.44 \\
\midrule
\multirow{3}{*}{Entity}
 & Whole & 73.89 & 79.03 & 63.72 & 69.43 & 78.02 & 60.59 \\
 & Part & 77.25 & 80.08 & 71.58 & 75.44 & 81.01 & 71.09 \\
 & State & 71.91 & 75.53 & 65.63 & 67.66 & 74.29 & 63.20 \\
\midrule
\multirow{5}{*}{Attribute}
 & Color & 87.90 & 88.52 & 84.71 & 85.32 & 87.87 & 82.35 \\
 & Shape & 78.93 & 80.57 & 72.93 & 79.59 & 80.57 & 78.17 \\
 & Size & 62.71 & 66.32 & 55.58 & 56.20 & 64.57 & 52.38 \\
 & Texture & 77.59 & 81.61 & 75.29 & 73.15 & 78.60 & 70.46 \\
 & Other & 77.47 & 80.43 & 72.39 & 73.00 & 77.72 & 69.26 \\
\midrule
\multirow{2}{*}{Relation}
 & Spatial & 86.10 & 89.17 & 80.72 & 82.82 & 87.46 & 77.67 \\
 & Non-spatial & 79.72 & 80.66 & 74.21 & 74.06 & 79.25 & 68.40 \\
\midrule
\multirow{2}{*}{Other}
 & Count & 53.88 & 60.25 & 43.38 & 49.63 & 62.00 & 39.13 \\
 & Text & 71.50 & 76.00 & 71.00 & 74.50 & 77.50 & 71.50 \\
\bottomrule
\end{tabular}
\end{table}

\subsection{GenEval Per-Task Breakdown}
\label{app:geneval-detail}

\Cref{tab:geneval-detail} reports the full per-task breakdown on GenEval~\citep{ghosh2024geneval} for all three methods at every guidance scale $s\in\{1,3,5,7,10\}$; the ``Overall'' column is the mean over the six tasks and reproduces the GenEval rows of \maincref{tab:cfg-scales}.
Focusing on the shared operating point $s{=}3$, SAGE improves four of the six categories, with the largest gain by far in color-attribute binding ($+11.5$\,pp), followed by two-object composition ($+3.5$\,pp) and single-object presence ($+3.4$\,pp); it shows small decreases in counting ($-1.9$\,pp) and colors ($-2.4$\,pp).
The gains therefore concentrate exactly on the compositional categories that depend on precise conditional control, which is where a write outside the conditional subspace is most damaging.
Across the whole sweep, SAGE leads the Baseline on the Overall score at every $s\ge 3$, while the KL routing constraint degrades every category at every scale, consistent with the loss of expert specialization discussed in \maincref{sec:exp-t2i-main}.

\begin{table}[h]
\centering
\caption{\textbf{GenEval per-task accuracy (\%) across CFG scales.} Per-category GenEval scores for all three methods at every guidance scale $s$; the ``Overall'' column is the mean over the six tasks and matches the GenEval rows of \maincref{tab:cfg-scales}.}
\label{tab:geneval-detail}
\small
\setlength{\tabcolsep}{4pt}
\begin{tabular}{@{}ll cccccc c@{}}
\toprule
Method & $s$ & Single Obj. & Two Obj. & Counting & Colors & Position & Color Attr. & Overall \\
\midrule
\multirow{5}{*}{Baseline}
 & 1  & 50.0 & 18.2 & 20.3 & 44.1 & 25.3 & 23.8 & 30.3 \\
 & 3  & 79.4 & 46.0 & 40.0 & 79.3 & 51.5 & 46.8 & 57.1 \\
 & 5  & 82.5 & 43.9 & 40.3 & 75.8 & 48.8 & 41.8 & 55.5 \\
 & 7  & 79.4 & 43.9 & 40.3 & 68.6 & 49.8 & 37.3 & 53.2 \\
 & 10 & 73.1 & 42.4 & 33.1 & 58.2 & 42.5 & 33.3 & 47.1 \\
\midrule
\multirow{5}{*}{SAGE ($\lambda{=}1$)}
 & 1  & 42.5 & 17.2 & 18.1 & 36.7 & 20.3 & 24.0 & 26.5 \\
 & 3  & 82.8 & 49.5 & 38.1 & 76.9 & 53.3 & 58.3 & 59.8 \\
 & 5  & 80.0 & 48.7 & 45.3 & 75.3 & 52.8 & 54.0 & 59.3 \\
 & 7  & 76.9 & 50.8 & 39.4 & 72.3 & 53.5 & 49.5 & 57.1 \\
 & 10 & 74.7 & 51.3 & 39.4 & 63.3 & 46.3 & 40.3 & 52.5 \\
\midrule
\multirow{5}{*}{KL ($\lambda_{\mathrm{kl}}{=}1$)}
 & 1  & 23.8 & 6.8  & 10.6 & 23.4 & 8.0  & 10.3 & 13.8 \\
 & 3  & 60.3 & 23.5 & 23.8 & 55.6 & 27.3 & 25.5 & 36.0 \\
 & 5  & 55.0 & 24.2 & 27.8 & 49.2 & 28.0 & 23.0 & 34.5 \\
 & 7  & 55.6 & 20.2 & 25.9 & 44.1 & 29.0 & 18.5 & 32.2 \\
 & 10 & 56.3 & 19.2 & 21.9 & 39.9 & 23.8 & 16.3 & 29.5 \\
\bottomrule
\end{tabular}
\end{table}

\subsection{Additional Ablations}
\label{app:ablations}

The two ablations below complement those reported in \maincref{sec:exp-ablation}, use the same setup as in \maincref{sec:exp-t2i-setup}, and are likewise evaluated at guidance scales $s\in\{3,5\}$. Both the activation-frequency ablation (\cref{tab:interval}) and the shared-expert and ProMoE~\citep{promoe2025} comparison (\cref{tab:shared-expert}) report DPG-Bench and GenEval scores at each scale.

\paragraph{Activation frequency.}
We investigate whether SAGE needs to be active at every training step or whether intermittent activation suffices.
Because $\Lsage$ is a minibatch subspace constraint rather than a
per-token identity, it can be amortized across steps: intermittent
application still reshapes how the two branches write. The table shows
that activating SAGE every second or every fifth step retains nearly
all of the from-scratch gain, which is consistent with a batch-level
section rather than a per-token constraint.
\Cref{tab:interval} compares activation at every step with activation at every second and every fifth step.

\begin{table}[h]
\centering
\caption{\textbf{Ablation: SAGE activation frequency ($s\in\{3,5\}$).} DPG-Bench and GenEval scores for SAGE activation at every step, every second step, and every fifth step.}
\label{tab:interval}
\small
\setlength{\tabcolsep}{6pt}
\renewcommand{\arraystretch}{1.15}
\begin{tabular}{@{}lcccc@{}}
\toprule
& \multicolumn{2}{c}{$s{=}3$} & \multicolumn{2}{c}{$s{=}5$} \\
\cmidrule(lr){2-3} \cmidrule(lr){4-5}
Interval & DPG & GenEval & DPG & GenEval \\
\midrule
Every step & 67.49 & 59.8 & 65.63 & 59.3 \\
Every 2 steps & 67.21 & 59.9 & 65.84 & 59.1 \\
Every 5 steps & 66.73 & 59.4 & 65.12 & 58.8 \\
\bottomrule
\end{tabular}
\end{table}

\paragraph{Shared expert as an implicit bridge.}
Two factors explain why existing MoE-DiTs~\citep{fei2024scaling} produce acceptable images despite the subspace leak.
The first is that moderate CFG hides the problem: at typical operating points ($s{=}3$ to $5$), the leak is present but not catastrophic, so the degradation without SAGE is tolerable, even though the Baseline still benefits substantially from SAGE at these scales ($+5.73$ DPG at $s{=}3$; \maincref{tab:cfg-scales}); the gap widens dramatically at $s{=}7$ to $10$, which matters increasingly as applications demand high guidance (double-CFG, style transfer, compositional prompting).
The second is that the shared expert is a dense-style path evaluated on both branches, which reduces exclusive routed mass (\cref{app:shared}) without putting $v_u$ in $\Vsc$.
We isolate this effect in \cref{tab:shared-expert} by removing the shared expert entirely and by comparing with ProMoE (which fixes one expert for the unconditional branch).

\begin{table}[h]
\centering
\caption{\textbf{Ablation: shared expert and ProMoE ($s\in\{3,5\}$).} Comparison of MoE configurations with and without SAGE against ProMoE at both guidance scales.}
\label{tab:shared-expert}
\small
\setlength{\tabcolsep}{4pt}
\begin{tabular}{@{}p{0.36\linewidth}cccc@{}}
\toprule
& \multicolumn{2}{c}{$s{=}3$} & \multicolumn{2}{c}{$s{=}5$} \\
\cmidrule(lr){2-3} \cmidrule(lr){4-5}
Configuration & DPG & GenEval & DPG & GenEval \\
\midrule
8E + 1 shared, no SAGE & 61.76 & 57.1 & 56.33 & 55.5 \\
8E + 1 shared, SAGE ($\lambda{=}1$) & 67.49 & 59.8 & 65.63 & 59.3 \\
8E, no shared, no SAGE & 57.82 & 53.0 & 52.41 & 51.4 \\
8E, no shared, SAGE ($\lambda{=}1$) & 64.05 & 56.4 & 62.18 & 55.6 \\
ProMoE & 63.12 & 58.4 & 57.95 & 56.8 \\
\bottomrule
\end{tabular}
\end{table}

The results show the expected pattern: removing the shared expert worsens baseline performance (more exclusive routed mass is exposed), while adding SAGE still recovers most of the gap.
ProMoE, which reserves one expert for the unconditional branch and thus plays a role analogous to the shared expert without aligning the remaining routed activations, provides partial relief but does not match SAGE.

\subsection{SAGE Preserves Routing Diversity}
\label{app:routing}

\begin{figure}[h]
\centering
\includegraphics[width=\linewidth]{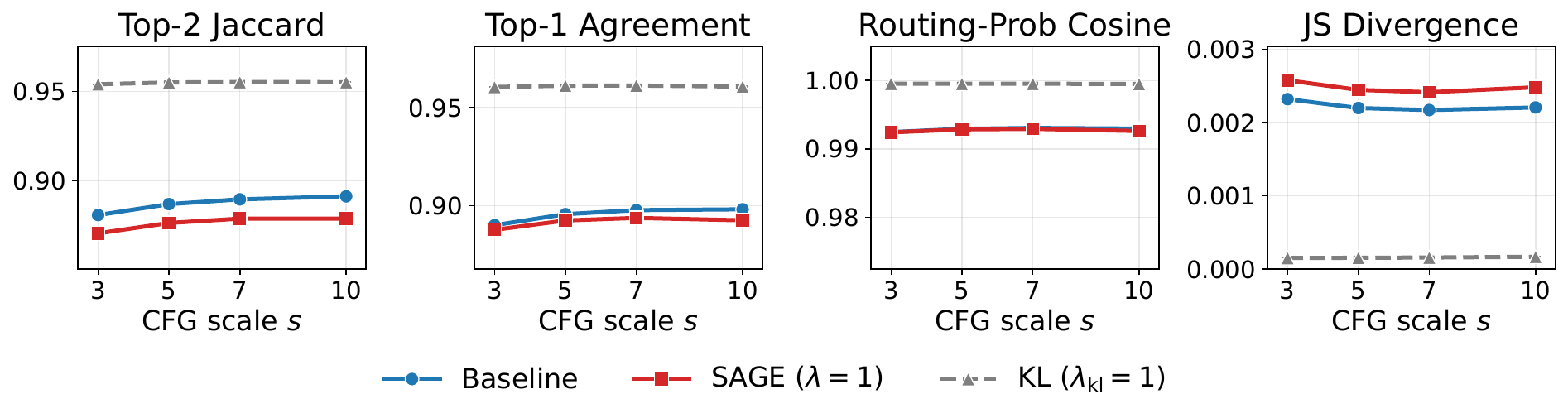}
\caption{\textbf{Conditional vs.\ unconditional router similarity across CFG scales.} SAGE (red) tracks the Baseline (blue) almost exactly on all four measures, whereas the KL routing constraint (gray, dashed) forces the two branches to agree (higher Jaccard/agreement, near-zero Jensen--Shannon (JS) divergence). Higher values indicate greater similarity for Top-$2$ Jaccard, Top-$1$ agreement, and routing-probability cosine similarity; lower values indicate greater similarity for JS divergence. Error bars are standard errors over $500$ prompts and are smaller than the markers.}
\label{fig:router-sim}
\end{figure}

\begin{figure}[h]
\centering
\includegraphics[width=0.8\linewidth]{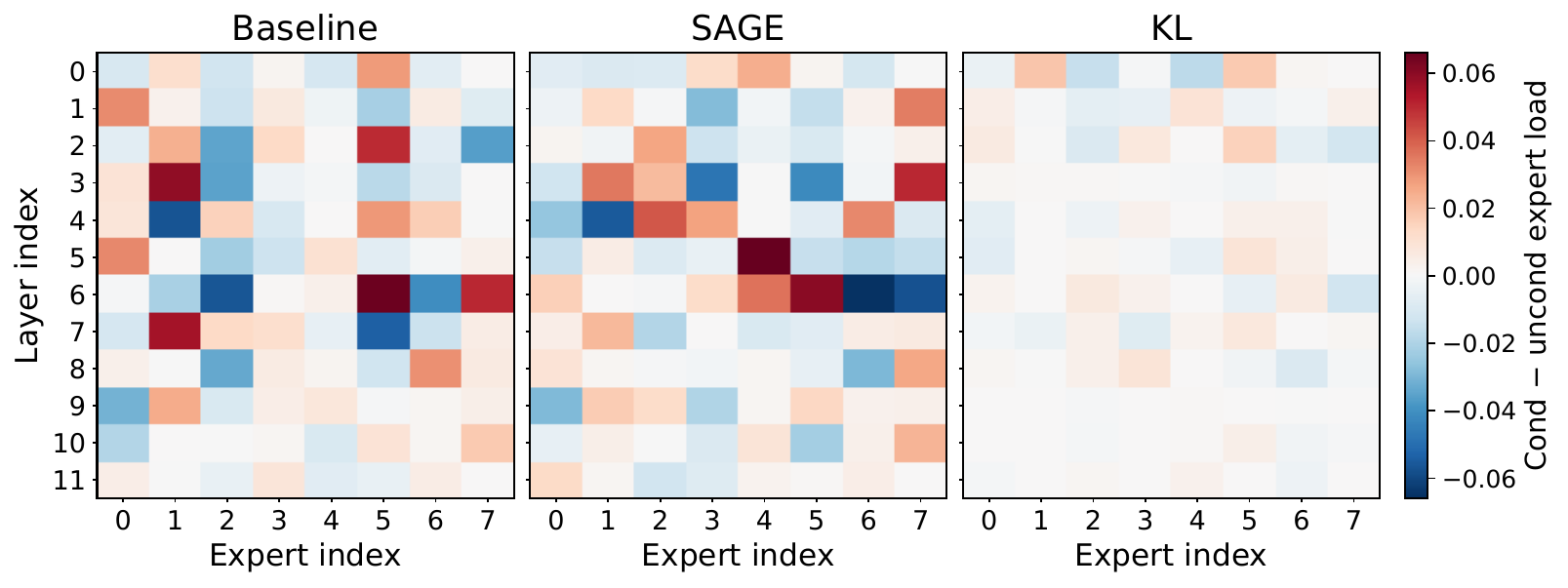}
\caption{\textbf{Conditional minus unconditional expert load per layer and expert at $s{=}3$}, averaged over $500$ prompts and shown on a shared color scale. Baseline and SAGE exhibit comparably large conditional/unconditional load differences (routing diversity preserved), whereas the KL constraint suppresses them almost entirely (routing forced to agree).}
\vspace{-0.2cm}
\label{fig:router-load}
\end{figure}

A central design goal of SAGE is to fix the subspace leak \emph{without} touching the router: unlike the KL routing constraint, which explicitly forces the two branches to route alike, SAGE only aligns unconditional activations to the conditional subspace (\maincref{sec:method-sage}).
Here we verify empirically that SAGE indeed leaves routing behavior essentially unchanged.
Using the 1B MoE-DiT of \maincref{sec:exp-t2i-setup}, we run each model on $500$ prompts across four guidance scales and record, at every denoising step and layer, the conditional and unconditional Top-$2$ expert selections and routing distributions. We evaluated various values for $\lambda_{\mathrm{kl}}$ (0.01, 0.1, and 1) and observed consistent results across all settings. Decreasing $\lambda_{\mathrm{kl}}$ did not yield improvements, as the routing health and benchmark scores still fell short of the baseline.

\paragraph{Conditional/unconditional routing similarity is unchanged.}
\Cref{fig:router-sim} reports four similarity measures between the conditional and unconditional branches as a function of the guidance scale.
Across all scales, SAGE is statistically indistinguishable from the Baseline on every measure: Top-$2$ Jaccard, Top-$1$ agreement, and routing-probability cosine similarity all coincide within noise, and the JS divergence between the two routing distributions is, if anything, marginally \emph{higher} for SAGE than for the Baseline, indicating that SAGE does not push the branches toward common experts.
The KL constraint behaves oppositely, driving Jaccard and agreement up toward $0.95$ and the JS divergence down toward zero, that is, collapsing the two branches onto nearly identical routes.

\paragraph{Per-expert load divergence is preserved.}
\Cref{fig:router-load} visualizes the per-layer, per-expert difference between the conditional and unconditional expert loads at $s{=}3$, on a shared color scale.
The Baseline and SAGE panels are equally vivid, with mean absolute load differences of $0.0144$ and $0.0148$, respectively, confirming that the two branches continue to recruit visibly different experts under SAGE.
By contrast, the KL panel is almost uniformly white (mean absolute difference $0.0034$, roughly $4\times$ smaller), showing that routing agreement has been enforced at the cost of expert specialization.
Together, the two figures confirm the claim of \maincref{sec:exp-t2i-main}: SAGE reduces the residual outside $\Vsc$ while leaving routing diversity intact, which is precisely why it avoids the specialization collapse that makes the KL constraint counterproductive.

\begin{table}[t]
\centering
\caption{\textbf{FID across CFG scales.} Lower is better.  $\Delta$ is Baseline $-$ SAGE, positive values indicate a SAGE improvement.}
\label{tab:fid-cfg}
\small
\setlength{\tabcolsep}{6pt}
\renewcommand{\arraystretch}{1.15}
\begin{tabular}{@{}l ccccc@{}}
\toprule
Method & $s{=}1$ & $s{=}3$ & $s{=}5$ & $s{=}7$ & $s{=}10$ \\
\midrule
Baseline            & 13.90 & 18.54 & 26.11 & 33.11 & 45.38 \\
SAGE ($\lambda{=}1$) & {13.18} & {14.80} & {20.28} & {25.54} & {37.12} \\
\midrule
$\Delta$ (Base $-$ SAGE) & $+0.72$ & $+3.74$ & $+5.83$ & $+7.57$ & $+8.26$ \\
\bottomrule
\end{tabular}
\vspace{-0.2cm}
\end{table}

\begin{figure}[t]
\vspace{-0.3cm}
\centering
\includegraphics[width=\linewidth]{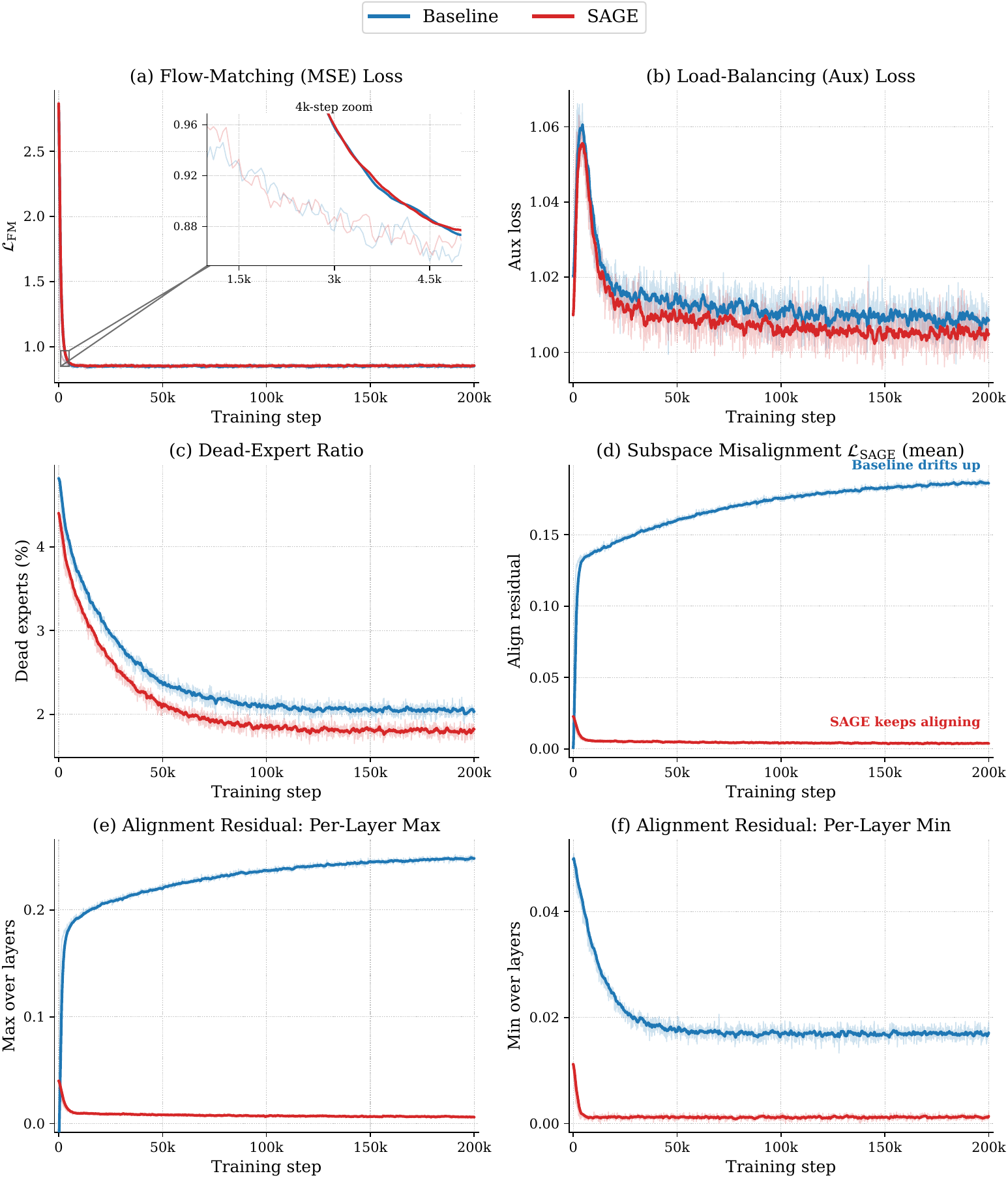}
\caption{\textbf{Training dynamics of the Baseline and SAGE.}
\textbf{(a)} Flow-matching MSE loss; the inset magnifies a $4{,}000$-step interval and shows that the two runs remain closely matched.
\textbf{(b)} MoE load-balancing auxiliary loss.
\textbf{(c)} Percentage of dead experts.
\textbf{(d--f)} Mean, per-layer maximum, and per-layer minimum of the subspace residual, respectively.
SAGE preserves the primary optimization and routing-health statistics while consistently reducing the alignment residual.}
\label{fig:training-dynamics}
\vspace{-0.3cm}
\end{figure}

\subsection{FID Across CFG Scales}
\label{app:fid}

To confirm that the CFG-robustness benefit of SAGE also holds for distributional image fidelity, we report the Fr\'echet Inception Distance (FID)~\citep{heusel2017fid} as a function of the guidance scale $s\in\{1,3,5,7,10\}$.

We sample a fixed set of $10{,}000$ prompts from the in-domain T2I test set and pair each prompt with its ground-truth image to form an in-domain reference set.
Reference images are decoded, EXIF-oriented, center-cropped to a square, and bicubic-resized to $320\times320$ to match the generation resolution; each model then generates one $320\times320$ image per prompt with $50$ sampling steps and a fixed per-prompt seed.
FID is computed from standard Inception-V3 features (2048-d \texttt{pool3}, bicubic resize to $299$), using the identical $10{,}000$ prompt IDs for the reference set and for every model and scale, so all conditions are strictly paired.
The Baseline and SAGE checkpoints are trained 200,000 steps.

\Cref{tab:fid-cfg} mirrors the trend of the alignment metrics.
At $s{=}1$ (no guidance) SAGE performs better than baseline ($13.90$ vs.\ $13.18$).
As the guidance scale grows, the Baseline FID degrades steeply ($+31.5$ from $s{=}1$ to $s{=}10$), whereas SAGE degrades markedly more slowly ($+23.94$); SAGE therefore attains a better FID at every tested scale.

\section{Training Dynamics and Health}
\label{app:training-dynamics}

We further compare the optimization dynamics of the Baseline and SAGE over a $200{,}000$-step training schedule. The purpose of this analysis is twofold: to verify that the additional alignment objective does not compromise the primary flow-matching objective or the health of MoE routing, and to confirm that it directly reduces the subspace residual targeted by SAGE.

\Cref{fig:training-dynamics}(a) reports the primary flow-matching objective $\Lfm$ defined in \maineqref{eq:fm-loss}. Both methods exhibit the same rapid initial decrease and converge to nearly identical values. Because the two curves overlap at the scale of the full plot, we additionally show a $4{,}000$-step magnified interval. The zoomed view confirms that their differences remain negligible.

 \Cref{fig:training-dynamics}(b) shows the standard MoE auxiliary loss used to encourage balanced expert utilization. A lower and stable value indicates that routing does not concentrate excessively on a small subset of experts. The two methods follow closely matched trajectories throughout training, with SAGE maintaining a slightly lower auxiliary loss than the Baseline. This indicates that SAGE does not interfere with the existing load-balancing mechanism or induce routing collapse.

Dead expert ratio is measured in \Cref{fig:training-dynamics}(c). Let $E$ denote the number of experts, $N$ the total number of routed tokens in the current batch, and $C_e$ the number of tokens assigned to expert $e$. We classify expert $e$ as dead when its load is below $10\%$ of the uniform expected load $N/E$, and compute
\begin{equation}
\mathrm{dead\_expert\_pct}
= \frac{1}{E}\sum_{e=1}^{E}
\mathbb{I}\!\left(C_e < 0.1\,\frac{N}{E}\right),
\label{eq:dead-expert-pct}
\end{equation}
where $\mathbb{I}(\cdot)$ is the indicator function. For both methods, the dead-expert percentage remains below $5\%$ and decreases as training proceeds. SAGE closely tracks the Baseline and does not increase the number of under-utilized experts, further confirming that its alignment constraint does not waste MoE capacity.

\Cref{fig:training-dynamics}(d--f) report three complementary summaries of the layer-wise alignment residual: its mean across MoE layers, its maximum, and its minimum. The Baseline residual grows over training in both the mean and worst-layer views, indicating that unconditional writes drift outside the conditional subspace $\Vsc$ as experts specialize. In contrast, SAGE rapidly decreases the residual and keeps its mean, maximum, and minimum close to zero throughout training. The per-layer maximum is especially informative because it rules out the possibility that a small set of poorly aligned layers is hidden by averaging.
\end{document}